%% file: cameraready.tex
\documentclass[twoside,11pt]{article}

\usepackage{jmlr2e}
\usepackage{amsmath}

\newcommand{\T}{\hat{t}}
\newcommand{\MaxPool}{\textrm{MaxPool}}

\newcommand{\floor}[1]{\lfloor #1 \rfloor}

\ShortHeadings{Multi-task Prediction of Disease Onsets from Longitudinal Lab Tests}{Razavian, Marcus, and Sontag}
\firstpageno{1}

\begin{document}

\title{Multi-task Prediction of Disease Onsets from Longitudinal Lab Tests}

\author{\center Narges Razavian, Jake Marcus, David Sontag \\ \normalfont Courant Institute of Mathematical Sciences, New York University \\ \{razavian,jmarcus,dsontag\}@cs.nyu.edu \\ \bigbreak}

\maketitle

\begin{abstract}
Disparate areas of machine learning have benefited from models that can take raw data with little preprocessing as input and learn rich representations of that raw data in order to perform well on a given prediction task. We evaluate this approach in healthcare by using longitudinal measurements of lab tests, one of the more raw signals of a patient's health state widely available in clinical data, to predict disease onsets. In particular, we train a Long Short-Term Memory (LSTM) recurrent neural network and two novel convolutional neural networks for multi-task prediction of disease onset for 133 conditions based on 18 common lab tests measured over time in a cohort of 298K patients derived from 8 years of administrative claims data. We compare the neural networks to a logistic regression with several hand-engineered, clinically relevant features. We find that the representation-based learning approaches significantly outperform this baseline. We believe that our work suggests a new avenue for patient risk stratification based solely on lab results. 
\end{abstract}

\section{Introduction}

The recent success of deep learning in disparate areas of machine learning has driven a shift towards machine learning models that can learn rich, hierarchical representations of raw data with little preprocessing and away from models that require manual construction of features by experts \citep{graves2005framewise, krizhevsky2012imagenet, mikolov2013distributed}. In natural language processing, for example, neural networks taking only character-level input achieve high performance on many tasks including text classification tasks \citep{zhang2015character, kimconvolutional}, machine translation \citep{ling2015character} and language modeling \citep{kim2016character}. 

Following these advances, attempts to learn features from raw medical signals have started to gain attention too. \citet{lasko2013computational} studied a method based on sparse auto-encoders to learn temporal variation features from 30-day uric acid observations to distinguish between gout and leukemia. \citet{che2015deep} developed a training method which, when datasets are small, allows prior domain knowledge to regularize the deeper layers of a feed-forward network for the task of multiple disease classification. Recent studies \citep{lipton2015learning, choi2015doctor} used Long Short-Term Memory (LSTM) recurrent neural networks (RNNs) for disease phenotyping. 

In this paper, we evaluate the representation-based learning approach in healthcare by using longitudinal measurements of laboratory tests, one of the more raw signals of a patient's health state widely available in clinical data, to predict disease onsets. We show that several multi-task neural networks, including a LSTM RNN and two novel convolutional neural networks, can aid in early diagnosis of a wide range of conditions (including conditions that the patient was not specifically tested for) without having to hand-engineer features for each condition. The source code of our implementation is available at \url{https://github.com/clinicalml/deepDiagnosis}.

\section{Prediction Task}\label{sec:res_pred}

\begin{figure}[t]
\begin{center}
\includegraphics[width=0.75\linewidth]{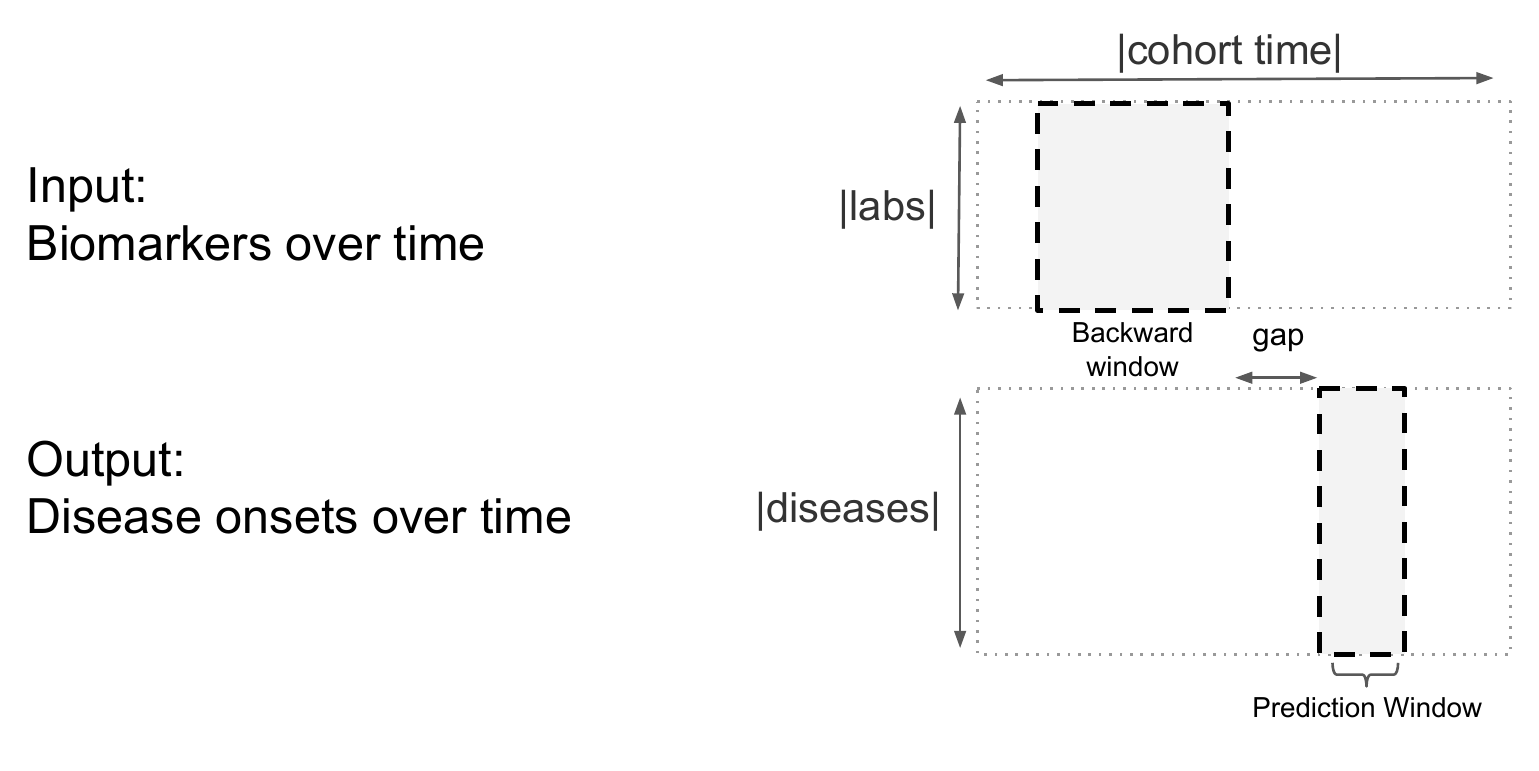}
\end{center}
\vspace{-7mm}
\caption{Overview of our prediction framework.}\label{fig:overview}
\end{figure}

Figure \ref{fig:overview} outlines the study's prediction framework. Our goal is early diagnosis of diseases for people who do not already have the disease. We required a 3-month gap between the end of the backward window, denoted $t$, and the start of the diagnosis window. The purpose of the 3 month gap was to ensure that the clinical tests taken right before the diagnosis of a disease would not allow our system to cheat in the prediction of that disease. Each output label was defined as positive if the diagnosis code for the disease was observed in at least $2$ distinct months between $3$ to $3+12$ months after $t$. Using $12$ months helps alleviate the noisy label problem. Requiring at least $2$ observations of the code also reduced the noise coming from physicians who report their wrong {\em suspected} diagnosis as a diagnosis. For each disease, we excluded individuals who already have the disease by time $t+3$. For exclusion, we required only $1$ diagnosis record instead of $2$ in order to remove patients who are even suspected of having the disease previously. This results in a more difficult, but more clinically meaningful prediction task.

Formally, we define the task of diagnosis as a supervised multi-task sequence classification problem. Each individual has a variable-length history of lab observations ($X$) and diagnosis records ($Y$). $X$ is continuous valued and $Y$ is binary. We use a sliding window framework to deal with variable length input. At each time point $t$ for each person, the model looks at a backward window of $B$ months of all $D$ biomarkers of the input, $X_{t-B:t}^{1:D}$, to predict the output. The output is a binary vector $Y$ of length $M$ indicating for each of the $M$ diseases whether they are newly diagnosed in the following months $t+3$ to $t + 3 + 12$, where $3$ is the {\em gap} and $12$ is the {\em prediction window}. 

\section{Cohort}

Our dataset consisted of lab measurement and diagnosis information for 298,000 individuals. The lab measurements had the resolution of 1 month and we used a backward window of 36 months for each prediction. These individuals were selected from a larger cohort of 4.1 million insurance subscribers between 2005 and 2013. We only included members who had at least one lab measurement per year for at least 3 consecutive years. 

We used lab tests that comprise a comprehensive metabolic panel plus cholesterol and bilirubin (18 lab tests in total), which are currently recommended annually and covered by most insurance companies in the United States. The names and codes of the labs used in our analysis are included in the Supplementary Materials. Each lab value was normalized by subtracting its mean and dividing by the standard deviation computed across the entire dataset. We randomly divided individuals into a 100K training set, a 100K validation set, and a 98K test set. The validation set was used to select the best epoch/parameters for models and prediction results are presented on the test set unseen during training and validation.

The predicted labels corresponded to diagnosis information for these individuals. In our dataset, each disease diagnosis is recorded as an ICD9-CM (International Classification of Diseases, Ninth Revision, Clinical Modification) code. 

\section{Methods} 

We now describe the baseline model, the two novel convolutional \citep{le1990handwritten,lecun1998gradient} architectures and the recurrent neural network with long short-term memory units \citep{hochreiter1997long} that we evaluate on this prediction task. The input to the baseline model are hand-engineered features derived from the patient's lab measurements, whereas the input to the representation-based models are the raw, sparse and asynchronously measured lab measurements. We also report the results of an ensemble of the representation-based models.

\subsection{Baseline}\label{sec:baseline}

We trained a Logistic Regression model on a large set of features derived from the patient's lab measurements. These features included the minimum, maximum and latest observation value for each of the labs as well as binary indicators for increasing and decreasing trends in the lab values within the backward window. The continuous features were computed on lab values that were normalized across the cohort (to have zero mean and unit variance). We used the validate data to choose the type and amount of L1, L2, and Dropout \citep{srivastava2014dropout} regularization, 
separately for each disease. 

\subsection{Multi-resolution Convolutional Neural Network (CNN1)}

\begin{figure*}[t]
\begin{center}
\includegraphics[width=\textwidth]{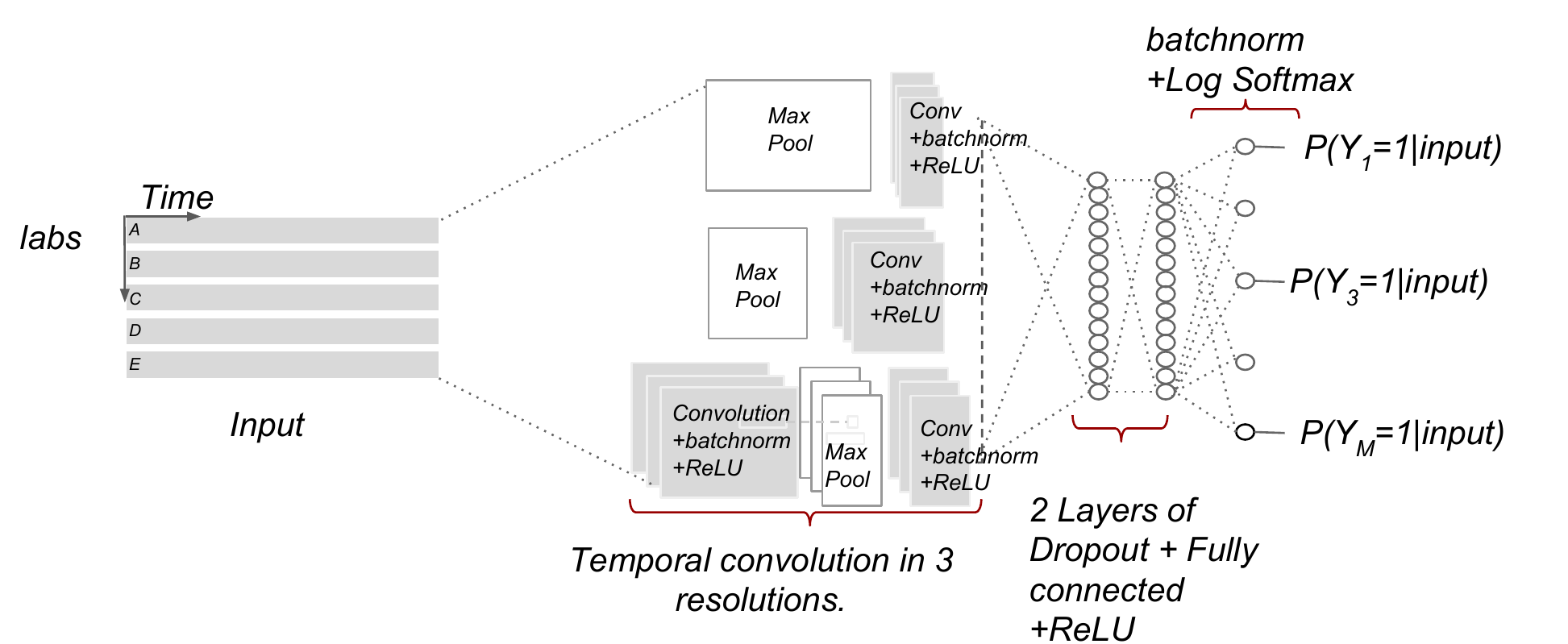}
\end{center}
\vspace{-2mm}
\caption{Architecture for Multi-resolution Convolutional Neural Network (CNN1)}\label{fig:predarch}
\end{figure*}

The architecture for our first convolutional neural network is shown in Figure \ref{fig:predarch}. We define $X_{t-B:t}^{1:D}$ to be the input to the network at time $t$ for $D$ lab measurements over the past $B$ months. Let $J$ be the number of filters in each convolution operator. Each filter $K_{i}^j$ ($j=1:J$) is of size $1 \times L$. The output of the convolution part of the network is a vector $C = [C_1, C_2, C_5]$ which is defined as follows:
\raggedbottom
\begin{align}
 C_1^{d,j} =& f(b_1^j + K_1^j * \MaxPool(X_{t-B:t}^{d}, p^2) ) \label{eq:maxpool2}\\ 
 C_2^{d,j} =& f(b_2^j + K_2^j * \MaxPool(X_{t-B:t}^{d}, p) ) \label{eq:maxpool}\\
 C_3^{d,j} =& f(b_3^j + K_3^j * X_{t-B:t}^{1:D}) \\
 C_4^{d,j} =& \MaxPool(C_3^{d,j}, p) \\
 C_5^{d,j} =& f(b_5^j + \sum_{k=1}^{J} K^j_5 * C_4^{d,k})
\end{align}

In the equations above, the nonlinearity $f$ is a rectified linear unit (ReLU) \citep{nair2010rectified} applied element-wise to a vector, and $*$ is the standard convolution operation.  $\MaxPool(Z,s)$ corresponds to a non-overlapping max pooling operation with step size $k$, defined as $\MaxPool(Z,s)[i] = \max(Z_{i\cdot s}, \ldots, Z_{(i+1)\cdot s -1})$ for $i = 1:\floor{\textrm{length}(Z)/s}$. We set $p=3$ for this paper. The vector $C_i$ is the concatenation of $C_i^{d,j}$ for all labs $d=1:D$ and filters $j=1:J$. The outputs of the first and second level in the multi-resolution network, ($C_1, C_2$), are results of the convolution operator applied to kernels $K_1^j$ and $K_2^j$ at different resolutions of the input. The third level of resolution includes two layers of convolution using filters $K_{3}^j$ and $K_{5}^j$. After every convolution operation, we use batch normalization \citep{ioffe2015batch}. 

After the multi-resolution convolution is applied, the vector $C= [C_1, C_2, C_5]$ represents the application of filters to all labs (note that the filters are shared across all the labs). We then use $2$ layers of hidden nodes to allow non-linear combination of filter activations on different labs:
\raggedbottom
\begin{align}
 h_1 = & f ( W_1^T C + b_{h_1}) \label{eq:hidden1}\\
 h_2 = & f ( W_2^T h_1 + b_{h_2})
\end{align} 

$W_i$ are the weights for the hidden nodes and $b_{h_i}$ is the bias associated with each layer. Each of the hidden layers are subject to Dropout regularization (with probability 0.5) during training, and are followed by batch normalization. 

Finally, for each disease $m=1:M$, the model predicts the likelihood of the disease via logistic regression over $h_2$:
\raggedbottom
\begin{align}
P(Y_{m} = 1 | X_{t-B:t}^{1:D}) = \sigma(W_m^T h_2 + b_m)\label{eq:logistic},
\end{align}
where $\sigma(x)=1/(1+e^{-x})$ is the sigmoid function. The loss function for each disease is the negative log likelihood of the true label, weighted by the inverse-label frequency to handle class imbalance during multi-task batch training. Diseases are trained independently, but the gradient is backpropagated through the shared part of the network for all diseases.

\subsection{Convolutional Neural Network over Time and Input dimensions (CNN2)}

\begin{figure*}[t]
\begin{center}
\includegraphics[width=1.03\textwidth]{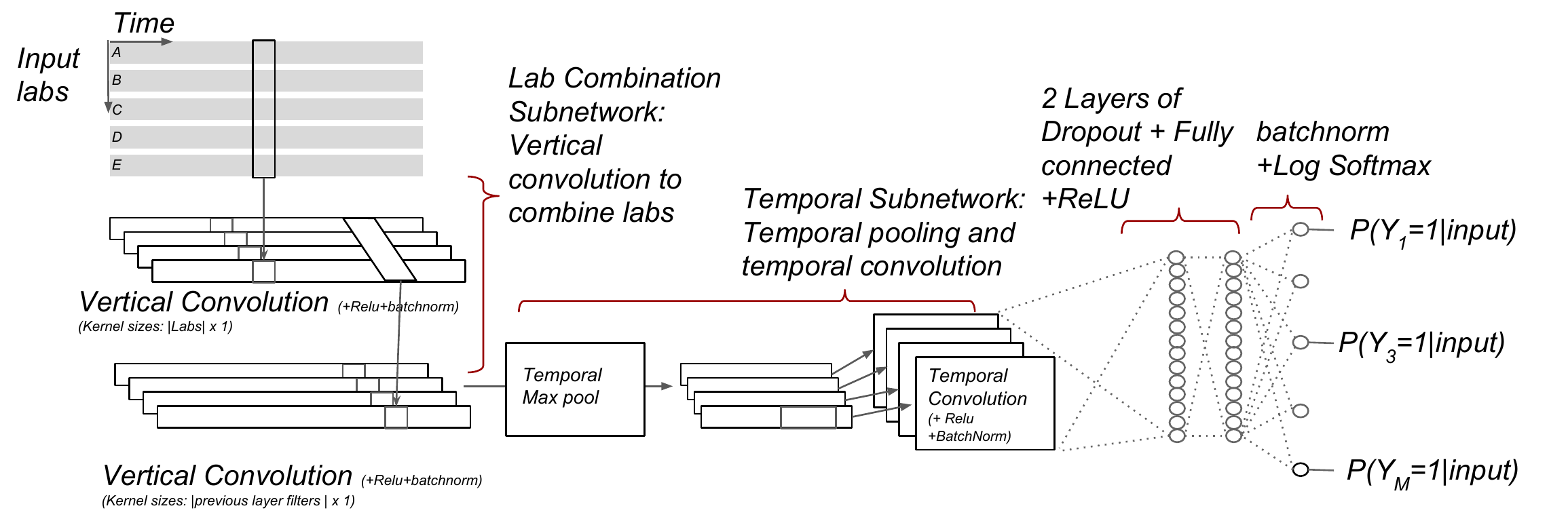}
\end{center}
\vspace{-4mm}
\caption{Architecture for Convolutional Neural Network over Time and Input dimensions (CNN2)}\label{fig:predarch2}
\end{figure*}

The architecture for our second convolutional neural network is shown in Figure \ref{fig:predarch2}. In this model, we first combine the labs via a vertical convolution with kernels that span across all labs. Having a few such combination layers enables us to project from the lab space into a new latent space which might better encode information about the labs. We then focus on temporal encoding of the result in the new space.

Given the input $X_{t-B:t}^{1:D}$, the output of the first vertical convolution with $L$ filters $K^{1:L}$, each of size ($D \times 1$), and nonlinearity $f$ is $V_{1:L}^{t-B:t}$ of size ($L \times 1 \times B$), where 
\raggedbottom
\begin{align}
 V_l^{t-B:t} = f(b_l + K^l * X_{t-B:t}^{1:D})
\end{align}
for $l=1:L$. We then repeat, applying new convolution filters of size $L \times 1$ to $V_{1:L}^{t-B:t}$, followed again by a nonlinearity $f$ (ReLU in our experiments), giving us two hidden layers in the vertical direction. Finally, temporal max pooling and convolution is applied to the last convolution output followed by two fully connected layers, similar to equations \eqref{eq:maxpool} and \eqref{eq:hidden1} through \eqref{eq:logistic}. Similar to the previous architecture, we optimize the weighted negative log-likelihood of the disease labels on the training data. 

\subsection{Long Short-Term Memory Network (LSTM)}

\begin{figure}[t]
\begin{center}
\includegraphics[width=0.5\linewidth]{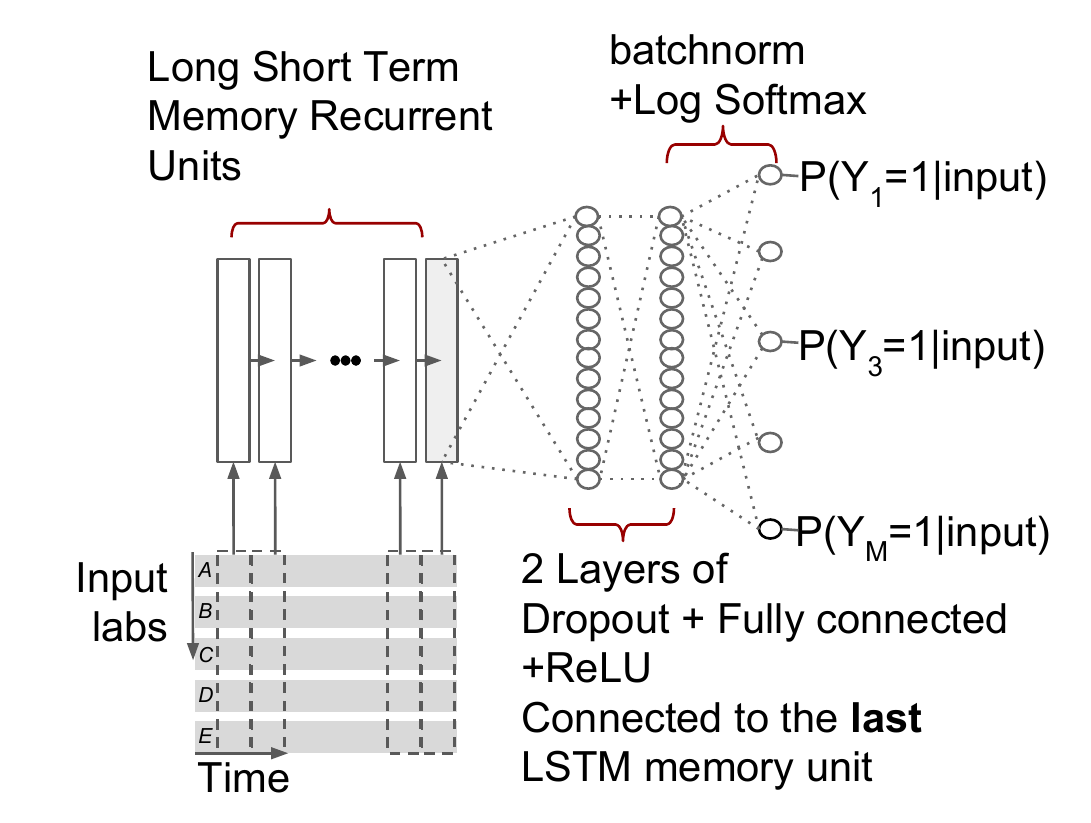}
\end{center}
\vspace{-7mm}
\caption{Architecture for the Long Short-Term Memory Network (LSTM)}\label{fig:lstm-last}
\end{figure}

The architecture for the Recurrent Neural Network with Long Short-Term Memory units \citep{hochreiter1997long} is shown in Figure \ref{fig:lstm-last}. These models encode a memory state $c_{\T}$ at each time step $\T$, which is only accessible through a particular gating mechanism. Given input $X_{\T}$ and the output and memory state of the recurrent network at time $\T-1$ ($h_{\T-1}$), the memory state and output for time steps $\hat{t}: t-B:t$ are computed as follows:
\raggedbottom
\begin{align}
i_{\T} =& \sigma(W_{x\rightarrow i} X_{\T} + W_{h\rightarrow i} h_{\T-1} + W_{c\rightarrow i}c_{\T-1} + b_{1\rightarrow i}) \\
f_{\T} =& \sigma(W_{x\rightarrow f} X_{\T} + W_{h\rightarrow f} h_{\T-1} + W_{c\rightarrow f}c_{\T-1} + b_{1\rightarrow f})   \\ 
z_{\T} =& \tanh(W_{x\rightarrow c} X_{\T} + W_{h\rightarrow c} h_{\T-1} + b_{1\rightarrow c})                  \\ 
c_{\T} =& f_{\T} c_{\T−1} + i_{\T} z_{\T}                                         \\
o_{\T} =& \sigma(W_{x\rightarrow o} X_{\T} + W_{h\rightarrow o} h_{\T-1} + W_{c\rightarrow o}c_{\T} + b_{1\rightarrow o})        \\
h_{\T} =& o_{\T} \tanh(c_{\T})
\end{align}
where $W_{*}$ and $b_{*}$ are the network's parameters, shared across all time steps. We use the output of the last time point, $h_{t}$, as the patient representation (i.e. $C$ in Eq. \eqref{eq:hidden1}). The rest of the network is the same as described in Eq. \eqref{eq:hidden1} through \eqref{eq:logistic} and we minimize the weighted negative log-likelihood.

\subsection{Weighted Batch Training to Deal with Class Imbalance}

We observed in our initial experiments that the predictive performance for more common diseases converges faster than for the uncommon diseases. Since early stopping is so important for preventing overfitting in neural networks, this leads to the following dilemma: either we stop early and underfit for the less common diseases, or we continue learning and overfit for the more common diseases. Decoupling them is not possible because of the shared patient representation. To alleviate this problem and following \citet{firat2016multi}, we use a weighted negative log-likelihood as the loss function. Specifically, we weight the gradient coming from each disease by the frequency of that disease. 
Our experiments indicated that the weighting improves the overall prediction results.

\section{Results} 

We used the validation set of 100K individuals to fine tune the hyperparameters of all our models. We then evaluated the best models on a test set of size 98K individuals. We describe the details of the architectures chosen in the Supplementary Materials. We report the Area Under the ROC curve (AUC) on the test set.  We implemented these experiments in Torch \citep{collobert2011torch7}. The source code of our implementation is available at \url{https://github.com/clinicalml/deepDiagnosis}.

Table \ref{tab:aucs} shows the AUC results for the top 25 diseases sorted by the maximum AUC that any model achieved on the test set. An ensemble of the neural networks performed best followed by the CNN2 architecture. The neural networks consistently outperformed the baseline in predicting the new onset of diseases 3 months in advance. In particular, heart failure, severe kidney diseases and liver problems, diabetes and hormone related conditions, and prostate cancer are among the diseases most accurately detected early from only 18 common lab measurements tracked over the previous 3 years. Our proposed models improve the quality of prediction for prostate cancer, elevated prostate specific antigen (note that the PSA lab is not part of our input), breast cancer, colon cancer, macular degeneration, and congestive heart failure most strongly. In the Supplementary Materials, we also report the top features from the baseline model for several of the diseases. 

\begin{table*}[t]
\begin{center}
\begin{tabular}{lllllll}
\multicolumn{1}{c}{ ICD9 Code and disease description	}&\multicolumn{1}{c}{\bf	LR	}&\multicolumn{1}{c}{\bf	LSTM	}&\multicolumn{1}{c}{\bf	CNN1	}&\multicolumn{1}{c}{\bf CNN2}&\multicolumn{1}{c}{\bf Ens}&\multicolumn{1}{c}{\bf Pos} \\
\hline 
585.6  End stage renal disease	&	0.886	&	0.917	&	0.910	&	0.916	&	0.920	&	837	\\
285.21 Anemia in chr kidney dis	&	0.849	&	0.866	&	0.868	&	0.880	&	0.879	&	1598	\\
585.3  Chr kidney dis stage III	&	0.846	&	0.851	&	0.857	&	0.858	&	0.864	&	2685	\\
584.9  Acute kidney failure NOS	&	0.805	&	0.820	&	0.828	&	0.831	&	0.835	&	3039	\\
250.01 DMI wo cmp nt st uncntrl	&	0.822	&	0.813	&	0.819	&	0.825	&	0.829	&	1522	\\
250.02 DMII wo cmp uncntrld	&	0.814	&	0.819	&	0.814	&	0.821	&	0.828	&	3519	\\
593.9  Renal and ureteral dis NOS	&	0.757	&	0.794	&	0.784	&	0.792	&	0.798	&	2111	\\
\bf428.0  CHF NOS	&	0.739	&	0.784	&	0.786	&	0.783	&	0.792	&	3479	\\
V053  Need prphyl vc vrl hepat	&	0.731	&	0.762	&	0.752	&	0.780	&	0.777	&	862	\\
\bf790.93 Elvtd prstate spcf antgn	&	0.666	&	0.758	&	0.761	&	0.768	&	0.772	&	1477	\\
\bf185   Malign neopl prostate	&	0.627	&	0.757	&	0.751	&	0.761	&	0.768	&	761	\\
274.9  Gout NOS	&	0.746	&	0.761	&	0.764	&	0.757	&	0.767	&	1529	\\
\bf362.52 Exudative macular degen	&	0.687	&	0.752	&	0.750	&	0.757	&	0.765	&	538	\\
\bf607.84 Impotence, organic orign	&	0.663	&	0.739	&	0.736	&	0.748	&	0.752	&	1372	\\
511.9  Pleural effusion NOS	&	0.708	&	0.736	&	0.742	&	0.746	&	0.749	&	2701	\\
\bf616.10 Vaginitis NOS	&	0.692	&	0.736	&	0.736	&	0.746	&	0.747	&	440	\\
\bf600.01 BPH w urinary obs/LUTS	&	0.648	&	0.737	&	0.737	&	0.738	&	0.747	&	1681	\\
\bf285.29 Anemia-other chronic dis	&	0.672	&	0.713	&	0.725	&	0.746	&	0.739	&	1075	\\
\bf346.90 Migrne unsp wo ntrc mgrn	&	0.633	&	0.736	&	0.710	&	0.724	&	0.732	&	471	\\
427.31 Atrial fibrillation	&	0.687	&	0.725	&	0.728	&	0.733	&	0.736	&	3766	\\
250.00 DMII wo cmp nt st uncntr	&	0.708	&	0.718	&	0.708	&	0.719	&	0.728	&	3125	\\
425.4  Prim cardiomyopathy NEC	&	0.683	&	0.718	&	0.719	&	0.722	&	0.726	&	1414	\\
728.87 Muscle weakness-general	&	0.683	&	0.704	&	0.718	&	0.722	&	0.723	&	4706	\\
\bf620.2  Ovarian cyst NEC/NOS	&	0.660	&	0.720	&	0.700	&	0.711	&	0.719	&	498	\\
286.9  Coagulat defect NEC/NOS	&	0.690	&	0.694	&	0.709	&	0.715	&	0.718	&	958	\\
\hline
\end{tabular}
\end{center}
\vspace{-2mm}
\caption{\small AUC results on the test set for different models for the top 25 diseases sorted by maximum AUC achieved by any of the models. Bold indicates that proposed models improve AUC by at least 0.05 compared to the baseline with hand-engineered features. Abbreviations: LR = Logistic Regression. CNN1 = Convolutional neural network architecture 1 (Figure \ref{fig:predarch}). CNN2 = Convolution neural network architecture 2 (Figure \ref{fig:predarch2}). LSTM = Long Short-Term Memory Network (Figure \ref{fig:lstm-last}). Ens = Ensemble of the deep models. Pos = Number of positive examples in the test set.}\label{tab:aucs}
\end{table*}

\section{Case study: Chronic Kidney Disease Progression}

We adapted the multitask architecture to predict the onset of end-stage renal disease (ESRD) requiring dialysis or a kidney transplant based on labs related to kidney function as well as diagnoses and prescriptions in a cohort of patients with advanced chronic kidney disease (CKD). 

Predictive models for ESRD in patients with advanced kidney disease could improve the timeliness of referral to a nephrologist enabling, for example, early counseling and education for high risk patients before they start dialysis \citep{green2012}. Clinical guidelines recommend that patients be referred to a nephrologist at least one year before they might be anticipated to require dialysis, and late referral may result in more rapid progression to kidney failure, worse quality of life for patients on dialysis, and missed opportunities for pre-emptive kidney transplantation \citep{ukrenalassociation2014}. 

See \citet{riskmodelreview} for a review of the literature on risk models for CKD. More recently, \citet{hagar2014} undertook a survival analysis for the progression of CKD using electronic health record data, \citet{perotte2015} developed a risk model to predict progression from Stage 3 to Stage 4 CKD, and \citet{fraccaro2016} evaluated several risk models for predicting the onset of CKD.   

\subsection{Data and Experiments Setup}

We use the same dataset as in the multi-disease prediction task. We restrict our analysis to patients with Stage 4 CKD, which we define as patients with at least 2 measurements of the estimated Glomerular Filtration Rate (eGFR) between 15 and 30 $\mathrm{mL/min/1.73m}^2$ observed at least 90 days apart \citep{kdigo2012}. We exclude patients with very sparse lab data by requiring at least one measurement of eGFR every 4 months of the training window. \cite{nice2014} recommends 2-3 measurements of eGFR a year for patients with Stage 4 CKD. 

We formulate the prediction task as taking a year of a patient's lab, diagnosis, prescription and demographic data as input and outputting a guess for whether or not that patient will start dialysis or undergo a kidney transplantation at any point in a 1-year window starting 3 months after the end of that year of clinical data. A training example for this prediction task consists of a matrix $X$ for a patient-year with $X$[$i$, $j$] = the value of the $i$th clinical or demographic feature (the average value for each lab, an indicator for each ICD9 code and drug class prescription, an indicator for gender and a continuous value for age) for the patient in the $j$th month of the year and an indicator $Y$ with $Y$ = 1 if the patient starts dialysis or undergoes a kidney transplantation in the 1-year outcome window and 0 otherwise. 

We included the labs associated with the most common LOINC codes for all of the labs used in the predictive models for kidney failure developed by \cite{tangri2011} and the labs with high prevalence in the CKD cohort analyzed by \cite{hagar2014}. We also included drug classes common in the treatment of kidney disease \citep{healthpartners2011} and ICD9 codes with high mutual information comparing positive to negative examples on the training data (withholding the validation and test data). Table \ref{tab:ckdfeatures} shows the final list of clinical features.

For each patient in the cohort, we obtain multiple training examples by constructing an $X$ for the one-year period starting at the 1st observation of eGFR for that patient, another $X$ for the one-year period starting at the 2nd observation of eGFR for that patient, and so on for every observation of eGFR in the patient's record. We exclude training examples where a dialysis CPT code appears before the start of the 1-year outcome window. 

This process results in 29,937 examples (5,484 patients) with 2,619 positive examples (781 patients).  We randomly divide these patients into 3 roughly equal groups and assign all the examples for a patient to the training, validation or test dataset.

Figure \ref{fig:ckdlabtimeseries} in the Supplementary Materials shows an example of lab data for a patient that does not start dialysis or undergo a kidney transplant in the outcome window and for a patient the starts dialysis in the outcome window.

We compared the performance of the CNN2 architecture adapted to this prediction task to two logistic regression baselines and a random forest. Additional details are provided in the Supplementary Materials. 

\subsection{Results}

The 4 models achieved similar performance on this prediction task (see Table \ref{tab:ckdauc}). The small sample size and the single binary outcome distinguish this task from the multi-disease setting and may make it difficult to observe large differences in performance between the models. A small number of features also seem to account for much of the signal. We observed that a logistic regression with a large L1 penalty achieves good performance on the task despite only using eGFR, urea nitrogen, age and gender as features.   

\section{Conclusion} 
In this work, we presented a large-scale application of two novel convolutional neural network architectures and a LSTM recurrent neural network for the task of multi-task early disease onset detection. These representation-based approaches significantly outperform a logistic regression with several hand-engineered, clinically relevant features. Interestingly, in our earlier work, we found that despite the large amount of missing data in the setting considered, preprocessing the data by imputing missing values did not significantly improve results \citep{razavian2015temporal}. As medical home and consumer healthcare technologies rapidly progress, we envision a growing role for automatic risk stratification of patients based solely on raw physiological and chemical signals.

\acks{The authors gratefully acknowledge support by Independence Blue Cross. The Tesla K40s used for
this research were donated by the NVIDIA Corporation. We thank Dr. Yindalon Aphinyanaphongs, Dr. Steven Horng, and Dr. Saul Blecker for providing helpful clinical perspectives throughout this research.}

\bibliography{cameraready}

\clearpage
\input{supp}

\end{document}

%% file: supp.tex
%

\section{Supplementary Materials \\ for Multi-task Prediction of Disease Onsets from Longitudinal Lab Tests}

\subsection{Cross-validation results}\label{sec:pred_spec}

For the convolutional models, we set the number of filters to be 64 for all the convolution layers with a kernel length of 8 (months) and a step size of 1. Each max-pooling module had a horizontal length of 3 and vertical length of 1 with a step size of 3 in the horizontal direction (i.e. no overlap). Each convolution module was followed by a batch normalization module (\cite{ioffe2015batch}) and then a ReLU nonlinearity (\cite{nair2010rectified}). We had 2 fully connected layers (with 100 nodes each, cross validated over [30,50, 100, 500,1000]) after concatenating the outputs of all the convolution layers. Each of the fully connected layers were followed by a batch normalization layer and a ReLU nonlinearity layer. We also added one Dropout module (\cite{srivastava2014dropout}) (0.5 dropout probability) before each fully connected layer. We tested models with and without batch-normalization and found that the networks converge much faster with batch-normalization. 

We had the following layers after the last ReLu nonlinearity for each of the 171 diseases: a Dropout layer(0.5 dropout probability), a fully connected layer (of size 2 nodes corresponding to binary outcome), a batch normalization layer and a Log Softmax Layer. A learning rate of $0.1$ was selected from among the values $[0.001, 0.01, 0.05, 0.1, 1]$ using the validation set average AUC (over all diseases) after 10 epochs. Training was done using Adadelta(\cite{zeiler2012adadelta}) optimization, which is a variant of stochastic gradient descent with adaptive step size. We used mini-batches of size 256. 

For the LSTM network, we cross-validated over the hidden LSTM units ([100 500 1000]), and 500 was selected as the best. For the shared part of the network, we used the best parameters found for the convolution models.

\subsection{Model details for CKD Case Study}

We compared the following models:

\begin{itemize}

\item CNN2. We applied the CNN2 architecture used in the multi-disease prediction task with 8 filters with kernel dimensions of 8x1. We used the raw clinical and demographic data as input without additional feature engineering. We chose the learning rate and the architecture based on cross-validation using random sampling of hyperparameters (\cite{bergstra2012}).

\item L2-regularized, logistic regression. For each lab, we added one feature for the average lab value across the training window. We included gender and the diagnosis and prescription data as binary indicators and age as a continuous variable. We chose a regularization constant based on cross-validation. 

\item L1-regularized, logistic regression. For each lab, we added features to the regression for the average lab value for the patient over the last 3 months of the training window, the past 6 months of the training window and over the entire training window. We also added binary features for whether or not the lab increased, decreased or fluctuated over the last 3 months, 6 months and over the entire training window. We included gender and the diagnosis and prescription data as binary indicators and age as a continuous variable. We chose a regularization constant based on cross-validation. 

\item Random forests. We used the raw clinical and demographic data as input without additional feature engineering. We chose the number of trees in the forest, the maximum depth of each tree, the maximum number of features to consider when looking for the best split, the minimum number of samples required to split a node, and the minimum number of samples in newly created leaves based on cross-validation using random sampling of hyperparameters. 

\end{itemize}

\subsection{Figures and Tables}

\begin{figure}[htbp]
\hspace*{-2cm} 
\includegraphics[scale=0.75]{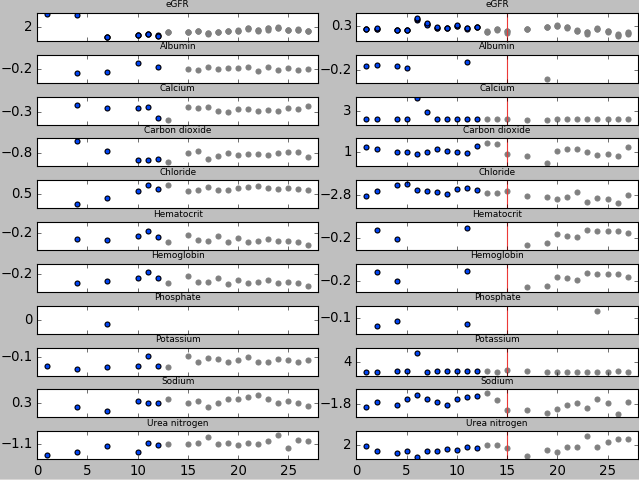}
\caption{Monthly average lab values for two patients. The left pane show the lab data for a patient who doesn't initiate dialysis or undergo a kidney transplant. The right pane shows a patient that starts dialysis in the outcome window. The vertical red line shows when that patient starts dialysis. The x-axis is the number of months from the beginning of the training window and the y-axis is the standardized lab value.}
\label{fig:ckdlabtimeseries} 
\end{figure}

\begin{table}[t]
\label{suppl_labs}
\caption{Name and LOINC of labs included as features for multi-task prediction}
\begin{center}
\begin{tabular}{lllll}
\multicolumn{1}{c}{\bf Lab name}&\multicolumn{1}{c}{\bf LOINC}
\\ \hline \\
Creatinine	& 2160-0 \\
Urea nitrogen	& 3094-0	\\
Potassium	& 2823-3 \\
Glucose	& 2345-7 \\
Alanine aminotransferase	& 1742-6 \\
Aspartate aminotransferase	& 1920-8	\\
Protein	& 2885-2 \\
Albumin	& 1751-7 \\
Cholesterol	& 2093-3	\\
Triglyceride	& 2571-8	\\
Cholesterol.in LDL	&	13457-7	\\
Calcium	&  17861-6 \\
Sodium	& 2951-2	\\
Chloride	& 2075-0	\\
Carbon dioxide	& 2028-9	\\
Urea nitrogen/Creatinine	&	3097-3	\\
Bilirubin	&	1975-2	\\
Albumin/Globulin	& 1759-0	\\
\end{tabular}
\end{center}
\end{table}

\begin{table*}[t]
\caption{Clinical features included in predictive models in the CKD case study}
\begin{center} \begin{tabular}{ll}
\multicolumn{1}{c}{Type}&\multicolumn{1}{c}{ Description } 
\\\hline \\
Lab & 33914-3 eGFR/1.73 sq M [Volume Rate/Area] in Serum or Plasma \\
Lab & 48642-3 eGFR/​1.73 sq M among non-blacks [Volume Rate/​Area] in Serum or Plasma \\
Lab & 48643-1 eGFR/​1.73 sq M among blacks [Volume Rate/​Area] in Serum or Plasma \\
Lab & 2160-0 Creatinine [Mass/​volume] in Serum or Plasma \\
Lab & 1751-7 Albumin [Mass/​volume] in Serum or Plasma \\
Lab & 17861-6 Calcium [Mass/​volume] in Serum or Plasma \\
Lab & 2028-9 Carbon dioxide, total [Moles/​volume] in Serum or Plasma \\
Lab & 9318-7 Albumin/​Creatinine [Mass Ratio] in Urine \\
Lab & 2777-1 Phosphate [Mass/​volume] in Serum or Plasma \\
Lab & 3094-0 Urea nitrogen [Mass/​volume] in Serum or Plasma \\
Lab & 2075-0 Chloride [Moles/​volume] in Serum or Plasma \\
Lab & 4544-3 Hematocrit [Volume Fraction] of Blood by Automated count \\
Lab & 718-7 Hemoglobin [Mass/​volume] in Blood \\
Lab & 2823-3 Potassium [Moles/​volume] in Serum or Plasma \\
Drug class & BETA-ADRENERGIC BLOCKING AGENTS \\
Drug class & LOOP DIURETICS \\
Drug class & HMG-COA REDUCTASE INHIBITORS \\
Drug class & DIHYDROPYRIDINES \\
Drug class & ANGIOTENSIN-CONVERTING ENZYME INHIBITORS \\
Drug class & ANGIOTENSIN II RECEPTOR ANTAGONISTS \\
Drug class & VITAMIN D \\
Drug class & DIRECT VASODILATORS \\
Drug class & THIAZIDE DIURETICS \\
Drug class & CHOLESTEROL ABSORPTION INHIBITORS \\
Drug class & THIAZIDE-LIKE DIURETICS \\
Drug class & PHOSPHATE-REMOVING AGENTS \\
Drug class & CENTRAL ALPHA-AGONISTS \\
Drug class & HEMATOPOIETIC AGENTS \\
Drug class & ALPHA-ADRENERGIC BLOCKING AGENTS \\
Diagnosis & 403.11 Ben hyp kid w cr kid V \\
Diagnosis & 403.91 Hyp kid NOS w cr kid V \\
Diagnosis & 285.21 Anemia in chr kidney dis \\
Diagnosis & 588.81 Sec hyperparathyrd-renal \\
Diagnosis & V72.81 Preop cardiovsclr exam \\
Diagnosis & 786.50 Chest pain NOS \\
Diagnosis & 600.00 BPH w/o urinary obs/LUTS \\
Diagnosis & 244.9 Hypothyroidism NOS \\
Diagnosis & 599.0 Urin tract infection NOS \\
Diagnosis & 250.02 DMII wo cmp uncntrld \\
Diagnosis & 250.01 DMI wo cmp nt st uncntrl \\
Diagnosis & 530.81 Esophageal reflux \\
Diagnosis & V58.61 Long-term use anticoagul \\
Diagnosis & 780.79 Malaise and fatigue NEC \\
Diagnosis & 562.10 Dvrtclo colon w/o hmrhg \\
\end{tabular} \end{center} 
\label{tab:ckdfeatures} 
\end{table*}

\begin{table*}[t]
\caption{ Area Under ROC curve for comparing the held out test score }
\begin{center} \begin{tabular}{ll}
\multicolumn{1}{c}{}&\multicolumn{1}{c}{ AUC } 
\\\hline \\
CNN2 & 0.774 \\
Random forest & 0.774  \\
L1-regularized logistic regression with hand-engineered features  & 0.768  \\
L2-regularized logistic regression  & 0.755  \\
\hline
\end{tabular} \end{center} 
\label{tab:ckdauc} 
\end{table*}

\input{top_features}

\clearpage

%% file: top_features.tex
\begin{table*}[t]
\caption{Top Features from the baseline model  for  585.6  End stage renal disease  }
\begin{center} \begin{tabular}{ll|ll}
\multicolumn{1}{c}{Feature}&\multicolumn{1}{c}{ weight }&\multicolumn{1}{c}{Feature}&\multicolumn{1}{c}{ weight }\\ \hline 
Glucose(2345-7)  -decreasing  & -1.099  & Alanine(1742-6)  -increasing  & 0.2263 \\
Chloride(2075-0)  -latest value  & 0.4988  & Cholesterol.in(13457-7)  -increasing  & -0.216 \\
Glucose(2345-7)  -latest value  & -0.485  & Urea(3094-0) -maximum  & -0.214 \\
Creatinine(2160-0) -maximum  & 0.4837  & Aspartate(1920-8) -maximum  & 0.2016 \\
Carbon(2028-9)  -increasing  & 0.4500  & Sodium(2951-2)  -latest value  & -0.191 \\
Cholesterol.in(13457-7) -minimum  & 0.3037  & Creatinine(2160-0)  -decreasing  & 0.1918 \\
Cholesterol.in(13457-7)  -latest value  & -0.254  & Chloride(2075-0) -maximum  & -0.177 \\
Glucose(2345-7) -minimum  & -0.251  & Carbon(2028-9) -minimum  & -0.170 \\
Calcium(17861-6)  -decreasing  & -0.239  & Alanine(1742-6) -minimum  & -0.122 \\
Urea(3094-0)  -increasing  & 0.2344  & Protein(2885-2)  -increasing  & 0.1208 \\
\end{tabular} \end{center} \end{table*}

\begin{table*}[t]
\caption{Top Features from the baseline model  for  285.21 Anemia in chr kidney dis  }
\begin{center} \begin{tabular}{ll|ll}
\multicolumn{1}{c}{Feature}&\multicolumn{1}{c}{ weight }&\multicolumn{1}{c}{Feature}&\multicolumn{1}{c}{ weight }\\ \hline 
Chloride(2075-0)  -latest value  & 0.7909  & Calcium(17861-6) -maximum  & -0.305 \\
Cholesterol.in(13457-7) -minimum  & 0.7244  & Triglyceride(2571-8)  -decreasing  & -0.301 \\
Glucose(2345-7)  -decreasing  & -0.717  & Triglyceride(2571-8)  -increasing  & -0.300 \\
Creatinine(2160-0) -maximum  & 0.5596  & Carbon(2028-9) -maximum  & -0.291 \\
Creatinine(2160-0) -minimum  & 0.4692  & Cholesterol.in(13457-7)  -increasing  & -0.290 \\
Chloride(2075-0)  -increasing  & -0.442  & Alanine(1742-6) -minimum  & -0.284 \\
Potassium(2823-3)  -increasing  & -0.381  & Glucose(2345-7)  -latest value  & -0.273 \\
Cholesterol.in(13457-7)  -latest value  & -0.370  & Alanine(1742-6)  -increasing  & 0.2531 \\
Aspartate(1920-8)  -decreasing  & 0.3658  & Carbon(2028-9)  -increasing  & 0.2383 \\
Glucose(2345-7) -minimum  & -0.324  & Urea(3094-0)  -increasing  & 0.2359 \\
\end{tabular} \end{center} \end{table*}
\clearpage

\begin{table*}[t]
\caption{Top Features from the baseline model  for  585.3  Chr kidney dis stage III  }
\begin{center} \begin{tabular}{ll|ll}
\multicolumn{1}{c}{Feature}&\multicolumn{1}{c}{ weight }&\multicolumn{1}{c}{Feature}&\multicolumn{1}{c}{ weight }\\ \hline 
Glucose(2345-7)  -decreasing  & -0.749  & Triglyceride(2571-8) -maximum  & 0.2169 \\
Cholesterol.in(13457-7) -minimum  & 0.6567  & Alanine(1742-6) -maximum  & 0.2060 \\
Chloride(2075-0)  -latest value  & 0.5660  & Cholesterol.in(13457-7)  -increasing  & -0.200 \\
Triglyceride(2571-8)  -increasing  & -0.425  & Potassium(2823-3)  -decreasing  & 0.1638 \\
Creatinine(2160-0) -maximum  & 0.4086  & Carbon(2028-9)  -increasing  & 0.1509 \\
Cholesterol.in(13457-7)  -latest value  & -0.374  & Alanine(1742-6) -minimum  & -0.143 \\
Chloride(2075-0) -maximum  & -0.361  & Glucose(2345-7) -minimum  & -0.140 \\
Creatinine(2160-0) -minimum  & 0.3368  & Calcium(17861-6)  -increasing  & 0.1407 \\
Glucose(2345-7)  -latest value  & -0.286  & Albumin(1751-7) -minimum  & 0.1385 \\
Alanine(1742-6)  -increasing  & 0.2667  & Potassium(2823-3) -minimum  & -0.137 \\
\end{tabular} \end{center} \end{table*}

\begin{table*}[t]
\caption{Top Features from the baseline model  for  584.9  Acute kidney failure NOS  }
\begin{center} \begin{tabular}{ll|ll}
\multicolumn{1}{c}{Feature}&\multicolumn{1}{c}{ weight }&\multicolumn{1}{c}{Feature}&\multicolumn{1}{c}{ weight }\\ \hline 
Glucose(2345-7)  -decreasing  & -0.729  & Carbon(2028-9)  -increasing  & 0.1915 \\
Creatinine(2160-0) -minimum  & 0.3994  & Alanine(1742-6) -minimum  & -0.180 \\
Glucose(2345-7)  -latest value  & -0.388  & Alanine(1742-6)  -increasing  & 0.1694 \\
Cholesterol.in(13457-7)  -latest value  & -0.372  & Triglyceride(2571-8)  -increasing  & -0.156 \\
Creatinine(2160-0) -maximum  & 0.3195  & Glucose(2345-7) -maximum  & 0.1402 \\
Chloride(2075-0)  -latest value  & 0.3087  & Potassium(2823-3) -minimum  & -0.136 \\
Creatinine(2160-0)  -decreasing  & 0.2750  & Chloride(2075-0)  -increasing  & -0.129 \\
Cholesterol.in(13457-7) -minimum  & 0.2571  & Aspartate(1920-8)  -latest value  & 0.1212 \\
Urea(3094-0)  -increasing  & 0.2374  & Potassium(2823-3)  -latest value  & 0.1078 \\
Albumin(1751-7) -maximum  & 0.2309  & Cholesterol.in(13457-7)  -increasing  & -0.105 \\
\end{tabular} \end{center} \end{table*}

\begin{table*}[t]
\caption{Top Features from the baseline model  for  250.01 DMI wo cmp nt st uncntrl  }
\begin{center} \begin{tabular}{ll|ll}
\multicolumn{1}{c}{Feature}&\multicolumn{1}{c}{ weight }&\multicolumn{1}{c}{Feature}&\multicolumn{1}{c}{ weight }\\ \hline 
Alanine(1742-6) -minimum  & -0.683  & Creatinine(2160-0) -minimum  & 0.1255 \\
Creatinine(2160-0)  -decreasing  & 0.6438  & Bilirubin(1975-2) -maximum  & 0.1244 \\
Glucose(2345-7)  -decreasing  & -0.235  & Aspartate(1920-8)  -increasing  & -0.116 \\
Glucose(2345-7) -minimum  & -0.227  & Cholesterol(2093-3)  -latest value  & 0.1157 \\
Urea(3094-0)  -decreasing  & 0.2204  & Alanine(1742-6)  -decreasing  & -0.092 \\
Sodium(2951-2)  -increasing  & 0.1755  & Urea(3097-3)  -latest value  & -0.078 \\
Urea(3094-0)  -increasing  & -0.172  & Protein(2885-2) -minimum  & 0.0750 \\
Protein(2885-2)  -increasing  & 0.1526  & Protein(2885-2)  -decreasing  & -0.072 \\
Albumin(1751-7) -maximum  & 0.1274  & Alanine(1742-6)  -latest value  & 0.0699 \\
Alanine(1742-6) -maximum  & -0.126  & Cholesterol.in(13457-7)  -increasing  & 0.0661 \\
\end{tabular} \end{center} \end{table*}
\clearpage

\begin{table*}[t]
\caption{Top Features from the baseline model  for  250.02 DMII wo cmp uncntrld  }
\begin{center} \begin{tabular}{ll|ll}
\multicolumn{1}{c}{Feature}&\multicolumn{1}{c}{ weight }&\multicolumn{1}{c}{Feature}&\multicolumn{1}{c}{ weight }\\ \hline 
Alanine(1742-6) -minimum  & -0.727  & Urea(3094-0) -minimum  & -0.165 \\
Creatinine(2160-0)  -decreasing  & 0.6069  & Aspartate(1920-8) -maximum  & 0.1647 \\
Cholesterol(2093-3)  -latest value  & 0.3880  & Albumin(1751-7) -maximum  & 0.1594 \\
Sodium(2951-2)  -increasing  & 0.3318  & Protein(2885-2)  -increasing  & 0.1521 \\
Alanine(1742-6)  -increasing  & -0.330  & Calcium(17861-6)  -increasing  & -0.146 \\
Urea(3094-0)  -decreasing  & 0.2875  & Triglyceride(2571-8)  -increasing  & -0.145 \\
Glucose(2345-7) -maximum  & 0.2016  & Albumin(1751-7) -minimum  & 0.1437 \\
Potassium(2823-3)  -latest value  & -0.184  & Cholesterol.in(13457-7) -minimum  & 0.1392 \\
Glucose(2345-7)  -decreasing  & -0.182  & Creatinine(2160-0)  -latest value  & -0.136 \\
Aspartate(1920-8)  -decreasing  & -0.168  & Alanine(1742-6) -maximum  & -0.136 \\
\end{tabular} \end{center} \end{table*}

\begin{table*}[t]
\caption{Top Features from the baseline model  for  593.9  Renal and ureteral dis NOS  }
\begin{center} \begin{tabular}{ll|ll}
\multicolumn{1}{c}{Feature}&\multicolumn{1}{c}{ weight }&\multicolumn{1}{c}{Feature}&\multicolumn{1}{c}{ weight }\\ \hline 
Glucose(2345-7)  -decreasing  & -0.769  & Urea(3094-0) -maximum  & -0.171 \\
Chloride(2075-0)  -latest value  & 0.5136  & Cholesterol.in(13457-7)  -increasing  & -0.150 \\
Cholesterol.in(13457-7) -minimum  & 0.4754  & Calcium(17861-6)  -decreasing  & 0.1446 \\
Cholesterol.in(13457-7)  -latest value  & -0.345  & Alanine(1742-6)  -decreasing  & -0.143 \\
Alanine(1742-6) -maximum  & 0.2568  & Sodium(2951-2)  -decreasing  & -0.142 \\
Creatinine(2160-0) -maximum  & 0.2394  & Carbon(2028-9) -minimum  & -0.137 \\
Creatinine(2160-0) -minimum  & 0.2163  & Cholesterol.in(13457-7)  -decreasing  & -0.133 \\
Creatinine(2160-0)  -decreasing  & 0.2093  & Carbon(2028-9)  -increasing  & 0.1204 \\
Calcium(17861-6) -maximum  & -0.182  & Albumin(1751-7) -maximum  & 0.1146 \\
Glucose(2345-7) -maximum  & 0.1816  & Alanine(1742-6) -minimum  & -0.113 \\
\end{tabular} \end{center} \end{table*}

\begin{table*}[t]
\caption{Top Features from the baseline model  for  428.0  CHF NOS  }
\begin{center} \begin{tabular}{ll|ll}
\multicolumn{1}{c}{Feature}&\multicolumn{1}{c}{ weight }&\multicolumn{1}{c}{Feature}&\multicolumn{1}{c}{ weight }\\ \hline 
Glucose(2345-7)  -decreasing  & -0.449  & Alanine(1742-6)  -increasing  & 0.1467 \\
Glucose(2345-7) -maximum  & 0.2493  & Chloride(2075-0)  -decreasing  & -0.146 \\
Cholesterol.in(13457-7) -minimum  & 0.2140  & Creatinine(2160-0) -maximum  & 0.1251 \\
Creatinine(2160-0)  -decreasing  & 0.2126  & Alanine(1742-6) -minimum  & -0.124 \\
Albumin(1751-7) -maximum  & 0.2045  & Creatinine(2160-0)  -latest value  & -0.120 \\
Chloride(2075-0)  -latest value  & 0.1996  & Albumin(1751-7)  -latest value  & -0.112 \\
Glucose(2345-7)  -latest value  & -0.195  & Aspartate(1920-8)  -decreasing  & 0.1106 \\
Creatinine(2160-0) -minimum  & 0.1911  & Cholesterol(2093-3)  -decreasing  & -0.100 \\
Calcium(17861-6)  -decreasing  & 0.1588  & Alanine(1742-6)  -decreasing  & 0.1000 \\
Cholesterol.in(13457-7) -maximum  & -0.156  & Alanine(1742-6)  -latest value  & -0.098 \\
\end{tabular} \end{center} \end{table*}
\clearpage

\begin{table*}[t]
\caption{Top Features from the baseline model  for  V05.3  Need prphyl vc vrl hepat  }
\begin{center} \begin{tabular}{ll|ll}
\multicolumn{1}{c}{Feature}&\multicolumn{1}{c}{ weight }&\multicolumn{1}{c}{Feature}&\multicolumn{1}{c}{ weight }\\ \hline 
Carbon(2028-9)  -increasing  & 0.4567  & Sodium(2951-2)  -latest value  & 0.0 \\
Creatinine(2160-0) -maximum  & 0.4441  & Sodium(2951-2)  -decreasing  & 0.0 \\
Glucose(2345-7)  -decreasing  & -0.233  & Sodium(2951-2)  -increasing  & 0.0 \\
Aspartate(1920-8)  -decreasing  & 0.1593  & Sodium(2951-2) -minimum  & 0.0 \\
Creatinine(2160-0) -minimum  & -0.119  & Calcium(17861-6)  -decreasing  & 0.0 \\
Urea(3094-0) -minimum  & -0.109  & Calcium(17861-6)  -latest value  & 0.0 \\
Creatinine(2160-0)  -increasing  & -0.058  & Calcium(17861-6)  -increasing  & 0.0 \\
Protein(2885-2) -maximum  & 0.0177  & Calcium(17861-6) -minimum  & 0.0 \\
Alanine(1742-6) -maximum  & 0.0164  & Calcium(17861-6) -maximum  & 0.0 \\
Chloride(2075-0) -maximum  & 0.0  & Cholesterol.in(13457-7)  -latest value  & 0.0 \\
\end{tabular} \end{center} \end{table*}

\begin{table*}[t]
\caption{Top Features from the baseline model  for  790.93 Elvtd prstate spcf antgn  }
\begin{center} \begin{tabular}{ll|ll}
\multicolumn{1}{c}{Feature}&\multicolumn{1}{c}{ weight }&\multicolumn{1}{c}{Feature}&\multicolumn{1}{c}{ weight }\\ \hline 
Alanine(1742-6) -maximum  & 0.2819  & Triglyceride(2571-8) -minimum  & 0.1261 \\
Potassium(2823-3) -minimum  & 0.2446  & Alanine(1742-6)  -increasing  & -0.120 \\
Sodium(2951-2)  -latest value  & 0.2335  & Protein(2885-2)  -increasing  & -0.115 \\
Cholesterol.in(13457-7) -minimum  & 0.2137  & Triglyceride(2571-8)  -increasing  & -0.096 \\
Creatinine(2160-0)  -decreasing  & -0.179  & Alanine(1742-6)  -decreasing  & 0.0861 \\
Cholesterol(2093-3)  -latest value  & 0.1770  & Sodium(2951-2)  -increasing  & 0.0832 \\
Potassium(2823-3)  -increasing  & -0.146  & Urea(3097-3)  -latest value  & 0.0789 \\
Glucose(2345-7) -minimum  & 0.1437  & Potassium(2823-3)  -decreasing  & -0.076 \\
Cholesterol(2093-3) -maximum  & -0.138  & Albumin(1751-7)  -decreasing  & -0.073 \\
Cholesterol(2093-3)  -increasing  & 0.1277  & Cholesterol(2093-3) -minimum  & -0.070 \\
\end{tabular} \end{center} \end{table*}

\begin{table*}[t]
\caption{Top Features from the baseline model  for  185.   Malign neopl prostate  }
\begin{center} \begin{tabular}{ll|ll}
\multicolumn{1}{c}{Feature}&\multicolumn{1}{c}{ weight }&\multicolumn{1}{c}{Feature}&\multicolumn{1}{c}{ weight }\\ \hline 
Cholesterol(2093-3) -minimum  & -0.293  & Cholesterol(2093-3)  -latest value  & 0.1599 \\
Aspartate(1920-8) -maximum  & 0.2527  & Sodium(2951-2)  -decreasing  & 0.1498 \\
Alanine(1742-6)  -latest value  & 0.2005  & Cholesterol(2093-3) -maximum  & -0.144 \\
Glucose(2345-7)  -decreasing  & -0.188  & Triglyceride(2571-8) -minimum  & 0.1375 \\
Cholesterol.in(13457-7) -minimum  & 0.1791  & Cholesterol(2093-3)  -increasing  & 0.1334 \\
Aspartate(1920-8)  -decreasing  & -0.174  & Alanine(1742-6)  -decreasing  & -0.130 \\
Alanine(1742-6)  -increasing  & -0.173  & Potassium(2823-3) -minimum  & 0.1305 \\
Potassium(2823-3)  -decreasing  & -0.168  & Potassium(2823-3) -maximum  & 0.1143 \\
Alanine(1742-6) -maximum  & 0.1676  & Urea(3094-0) -maximum  & -0.110 \\
Sodium(2951-2)  -increasing  & 0.1648  & Bilirubin(1975-2)  -decreasing  & -0.108 \\
\end{tabular} \end{center} \end{table*}
\clearpage

\begin{table*}[t]
\caption{Top Features from the baseline model  for  274.9  Gout NOS  }
\begin{center} \begin{tabular}{ll|ll}
\multicolumn{1}{c}{Feature}&\multicolumn{1}{c}{ weight }&\multicolumn{1}{c}{Feature}&\multicolumn{1}{c}{ weight }\\ \hline 
Glucose(2345-7)  -decreasing  & -0.414  & Creatinine(2160-0)  -decreasing  & 0.1201 \\
Cholesterol.in(13457-7)  -latest value  & -0.266  & Urea(3094-0)  -increasing  & 0.1181 \\
Aspartate(1920-8)  -increasing  & -0.261  & Carbon(2028-9)  -increasing  & 0.1005 \\
Cholesterol.in(13457-7) -minimum  & 0.2584  & Aspartate(1920-8) -maximum  & 0.0980 \\
Creatinine(2160-0) -maximum  & 0.2092  & Protein(2885-2)  -decreasing  & -0.096 \\
Glucose(2345-7)  -latest value  & -0.200  & Chloride(2075-0)  -latest value  & 0.0953 \\
Glucose(2345-7)  -increasing  & -0.194  & Creatinine(2160-0) -minimum  & 0.0881 \\
Urea(3094-0) -minimum  & 0.1545  & Creatinine(2160-0)  -latest value  & -0.086 \\
Alanine(1742-6) -maximum  & 0.1316  & Aspartate(1920-8)  -latest value  & -0.078 \\
Protein(2885-2)  -latest value  & 0.1295  & Cholesterol(2093-3)  -increasing  & -0.077 \\
\end{tabular} \end{center} \end{table*}

\begin{table*}[t]
\caption{Top Features from the baseline model  for  362.52 Exudative macular degen  }
\begin{center} \begin{tabular}{ll|ll}
\multicolumn{1}{c}{Feature}&\multicolumn{1}{c}{ weight }&\multicolumn{1}{c}{Feature}&\multicolumn{1}{c}{ weight }\\ \hline 
Creatinine(2160-0) -maximum  & 0.3761  & Glucose(2345-7)  -decreasing  & 0.1683 \\
Creatinine(2160-0)  -latest value  & -0.270  & Creatinine(2160-0) -minimum  & 0.1545 \\
Alanine(1742-6)  -latest value  & -0.254  & Cholesterol.in(13457-7)  -increasing  & 0.1477 \\
Urea(3094-0)  -increasing  & -0.241  & Albumin(1751-7)  -latest value  & -0.145 \\
Alanine(1742-6)  -increasing  & 0.2399  & Potassium(2823-3)  -latest value  & 0.1365 \\
Alanine(1742-6) -minimum  & 0.2304  & Aspartate(1920-8)  -increasing  & -0.120 \\
Glucose(2345-7)  -latest value  & -0.200  & Calcium(17861-6)  -decreasing  & 0.1134 \\
Potassium(2823-3) -minimum  & 0.1818  & Glucose(2345-7) -minimum  & 0.0982 \\
Cholesterol.in(13457-7)  -latest value  & 0.1795  & Aspartate(1920-8) -minimum  & 0.0966 \\
Urea(3094-0) -maximum  & -0.168  & Carbon(2028-9)  -increasing  & -0.090 \\
\end{tabular} \end{center} \end{table*}

\begin{table*}[t]
\caption{Top Features from the baseline model  for  607.84 Impotence, organic orign  }
\begin{center} \begin{tabular}{ll|ll}
\multicolumn{1}{c}{Feature}&\multicolumn{1}{c}{ weight }&\multicolumn{1}{c}{Feature}&\multicolumn{1}{c}{ weight }\\ \hline 
Cholesterol(2093-3) -minimum  & -0.296  & Urea(3094-0)  -increasing  & -0.140 \\
Alanine(1742-6) -maximum  & 0.2400  & Glucose(2345-7)  -latest value  & 0.1306 \\
Alanine(1742-6)  -increasing  & -0.237  & Aspartate(1920-8)  -decreasing  & -0.116 \\
Aspartate(1920-8)  -increasing  & -0.210  & Urea(3094-0) -minimum  & -0.111 \\
Potassium(2823-3)  -increasing  & 0.1967  & Protein(2885-2) -maximum  & 0.1038 \\
Potassium(2823-3)  -decreasing  & 0.1671  & Chloride(2075-0) -minimum  & 0.0963 \\
Creatinine(2160-0) -maximum  & -0.160  & Cholesterol.in(13457-7) -minimum  & 0.0905 \\
Protein(2885-2)  -latest value  & -0.153  & Creatinine(2160-0)  -increasing  & -0.086 \\
Creatinine(2160-0)  -decreasing  & 0.1531  & Aspartate(1920-8)  -latest value  & 0.0856 \\
Calcium(17861-6)  -latest value  & -0.141  & Alanine(1742-6)  -decreasing  & -0.078 \\
\end{tabular} \end{center} \end{table*}
\clearpage

\begin{table*}[t]
\caption{Top Features from the baseline model  for  511.9  Pleural effusion NOS  }
\begin{center} \begin{tabular}{ll|ll}
\multicolumn{1}{c}{Feature}&\multicolumn{1}{c}{ weight }&\multicolumn{1}{c}{Feature}&\multicolumn{1}{c}{ weight }\\ \hline 
Alanine(1742-6)  -increasing  & 0.4040  & Triglyceride(2571-8)  -increasing  & -0.194 \\
Glucose(2345-7)  -decreasing  & -0.366  & Creatinine(2160-0) -maximum  & 0.1678 \\
Urea(3094-0)  -latest value  & -0.297  & Aspartate(1920-8)  -increasing  & -0.166 \\
Chloride(2075-0)  -latest value  & 0.2698  & Cholesterol.in(13457-7)  -decreasing  & -0.158 \\
Urea(3094-0)  -increasing  & 0.2680  & Glucose(2345-7)  -latest value  & -0.154 \\
Chloride(2075-0) -maximum  & -0.251  & Cholesterol(2093-3)  -decreasing  & -0.154 \\
Aspartate(1920-8)  -latest value  & 0.2400  & Creatinine(2160-0)  -latest value  & 0.1370 \\
Cholesterol.in(13457-7) -minimum  & 0.2396  & Cholesterol.in(13457-7) -maximum  & -0.135 \\
Alanine(1742-6) -minimum  & -0.223  & Albumin(1751-7) -maximum  & 0.1317 \\
Alanine(1742-6) -maximum  & -0.200  & Triglyceride(2571-8) -maximum  & -0.129 \\
\end{tabular} \end{center} \end{table*}

\begin{table*}[t]
\caption{Top Features from the baseline model  for  616.10 Vaginitis NOS  }
\begin{center} \begin{tabular}{ll|ll}
\multicolumn{1}{c}{Feature}&\multicolumn{1}{c}{ weight }&\multicolumn{1}{c}{Feature}&\multicolumn{1}{c}{ weight }\\ \hline 
Alanine(1742-6) -maximum  & -0.567  & Cholesterol(2093-3) -maximum  & 0.1796 \\
Protein(2885-2)  -decreasing  & -0.386  & Creatinine(2160-0) -maximum  & -0.172 \\
Alanine(1742-6)  -increasing  & 0.3732  & Cholesterol.in(13457-7) -minimum  & -0.151 \\
Creatinine(2160-0)  -increasing  & -0.340  & Calcium(17861-6)  -decreasing  & 0.1461 \\
Calcium(17861-6) -maximum  & 0.2638  & Triglyceride(2571-8)  -increasing  & 0.0902 \\
Urea(3094-0)  -latest value  & -0.263  & Bilirubin(1975-2) -minimum  & -0.075 \\
Creatinine(2160-0) -minimum  & -0.230  & Albumin/Globulin(1759-0)  -increasing  & 0.0708 \\
Protein(2885-2) -minimum  & -0.200  & Urea(3094-0)  -increasing  & -0.067 \\
Cholesterol.in(13457-7)  -latest value  & 0.1909  & Glucose(2345-7)  -decreasing  & 0.0619 \\
Glucose(2345-7) -minimum  & -0.190  & Potassium(2823-3)  -decreasing  & -0.058 \\
\end{tabular} \end{center} \end{table*}

\begin{table*}[t]
\caption{Top Features from the baseline model  for  600.01 BPH w urinary obs/LUTS  }
\begin{center} \begin{tabular}{ll|ll}
\multicolumn{1}{c}{Feature}&\multicolumn{1}{c}{ weight }&\multicolumn{1}{c}{Feature}&\multicolumn{1}{c}{ weight }\\ \hline 
Glucose(2345-7)  -decreasing  & -0.358  & Sodium(2951-2)  -latest value  & 0.1253 \\
Alanine(1742-6)  -increasing  & -0.324  & Chloride(2075-0)  -latest value  & 0.1236 \\
Creatinine(2160-0)  -latest value  & -0.255  & Cholesterol.in(13457-7)  -decreasing  & -0.114 \\
Protein(2885-2)  -latest value  & -0.253  & Aspartate(1920-8)  -increasing  & -0.100 \\
Sodium(2951-2)  -decreasing  & 0.2298  & Creatinine(2160-0)  -decreasing  & 0.0918 \\
Cholesterol.in(13457-7) -minimum  & 0.1959  & Cholesterol.in(13457-7)  -latest value  & -0.091 \\
Aspartate(1920-8) -maximum  & 0.1736  & Chloride(2075-0)  -increasing  & -0.083 \\
Glucose(2345-7)  -increasing  & -0.134  & Urea(3094-0) -maximum  & -0.076 \\
Creatinine(2160-0) -maximum  & -0.133  & Potassium(2823-3)  -increasing  & -0.072 \\
Urea(3094-0)  -latest value  & 0.1271  & Chloride(2075-0)  -decreasing  & -0.071 \\
\end{tabular} \end{center} \end{table*}
\clearpage

\begin{table*}[t]
\caption{Top Features from the baseline model  for  285.29 Anemia-other chronic dis  }
\begin{center} \begin{tabular}{ll|ll}
\multicolumn{1}{c}{Feature}&\multicolumn{1}{c}{ weight }&\multicolumn{1}{c}{Feature}&\multicolumn{1}{c}{ weight }\\ \hline 
Glucose(2345-7)  -decreasing  & -0.492  & Triglyceride(2571-8)  -increasing  & -0.205 \\
Creatinine(2160-0)  -increasing  & 0.3508  & Urea(3094-0) -maximum  & -0.189 \\
Protein(2885-2)  -latest value  & 0.3098  & Albumin(1751-7) -maximum  & 0.1885 \\
Aspartate(1920-8)  -decreasing  & 0.3042  & Aspartate(1920-8) -maximum  & -0.183 \\
Carbon(2028-9)  -increasing  & 0.2610  & Chloride(2075-0)  -latest value  & 0.1796 \\
Creatinine(2160-0) -maximum  & 0.2594  & Potassium(2823-3)  -latest value  & -0.168 \\
Alanine(1742-6)  -increasing  & 0.2468  & Aspartate(1920-8)  -latest value  & 0.1632 \\
Creatinine(2160-0) -minimum  & 0.2386  & Creatinine(2160-0)  -decreasing  & 0.1492 \\
Potassium(2823-3) -minimum  & 0.2368  & Cholesterol.in(13457-7)  -decreasing  & -0.147 \\
Cholesterol.in(13457-7)  -increasing  & -0.216  & Cholesterol(2093-3)  -latest value  & -0.134 \\
\end{tabular} \end{center} \end{table*}

\begin{table*}[t]
\caption{Top Features from the baseline model  for  346.90 Migrne unsp wo ntrc mgrn  }
\begin{center} \begin{tabular}{ll|ll}
\multicolumn{1}{c}{Feature}&\multicolumn{1}{c}{ weight }&\multicolumn{1}{c}{Feature}&\multicolumn{1}{c}{ weight }\\ \hline 
Cholesterol.in(13457-7) -minimum  & -0.403  & Aspartate(1920-8) -maximum  & -0.105 \\
Cholesterol(2093-3)  -increasing  & 0.3081  & Alanine(1742-6)  -latest value  & -0.102 \\
Cholesterol.in(13457-7)  -latest value  & 0.2965  & Creatinine(2160-0)  -increasing  & -0.096 \\
Alanine(1742-6) -minimum  & 0.2748  & Protein(2885-2)  -decreasing  & 0.0964 \\
Sodium(2951-2)  -decreasing  & -0.204  & Sodium(2951-2) -minimum  & 0.0925 \\
Cholesterol.in(13457-7)  -decreasing  & 0.1985  & Potassium(2823-3)  -latest value  & -0.088 \\
Aspartate(1920-8)  -increasing  & 0.1974  & Potassium(2823-3)  -decreasing  & -0.084 \\
Albumin/Globulin(1759-0)  -increasing  & 0.1783  & Carbon(2028-9) -maximum  & 0.0836 \\
Glucose(2345-7)  -latest value  & -0.132  & Albumin/Globulin(1759-0) -maximum  & 0.0587 \\
Protein(2885-2) -maximum  & 0.1100  & Calcium(17861-6)  -decreasing  & 0.0528 \\
\end{tabular} \end{center} \end{table*}

\begin{table*}[t]
\caption{Top Features from the baseline model  for  427.31 Atrial fibrillation  }
\begin{center} \begin{tabular}{ll|ll}
\multicolumn{1}{c}{Feature}&\multicolumn{1}{c}{ weight }&\multicolumn{1}{c}{Feature}&\multicolumn{1}{c}{ weight }\\ \hline 
Creatinine(2160-0) -maximum  & 0.2260  & Albumin/Globulin(1759-0)  -latest value  & -0.019 \\
Alanine(1742-6)  -increasing  & 0.1981  & Creatinine(2160-0)  -latest value  & -0.015 \\
Glucose(2345-7) -maximum  & 0.1956  & Urea(3097-3)  -latest value  & -0.011 \\
Glucose(2345-7)  -latest value  & -0.099  & Sodium(2951-2)  -decreasing  & 0.0089 \\
Urea(3094-0) -maximum  & -0.089  & Creatinine(2160-0) -minimum  & 0.0084 \\
Urea(3094-0)  -latest value  & -0.082  & Cholesterol.in(13457-7)  -latest value  & -0.007 \\
Glucose(2345-7)  -decreasing  & -0.079  & Cholesterol(2093-3)  -decreasing  & -0.000 \\
Alanine(1742-6) -minimum  & -0.066  & Potassium(2823-3) -minimum  & 0.0 \\
Urea(3094-0) -minimum  & -0.037  & Chloride(2075-0) -minimum  & 0.0 \\
Cholesterol(2093-3)  -increasing  & -0.031  & Chloride(2075-0) -maximum  & 0.0 \\
\end{tabular} \end{center} \end{table*}
\clearpage

\begin{table*}[t]
\caption{Top Features from the baseline model  for  250.00 DMII wo cmp nt st uncntr  }
\begin{center} \begin{tabular}{ll|ll}
\multicolumn{1}{c}{Feature}&\multicolumn{1}{c}{ weight }&\multicolumn{1}{c}{Feature}&\multicolumn{1}{c}{ weight }\\ \hline 
Alanine(1742-6) -minimum  & -0.406  & Sodium(2951-2) -maximum  & -0.104 \\
Aspartate(1920-8)  -increasing  & -0.371  & Potassium(2823-3)  -latest value  & -0.099 \\
Creatinine(2160-0)  -decreasing  & 0.2839  & Urea(3097-3) -maximum  & 0.0990 \\
Albumin(1751-7) -maximum  & 0.2038  & Triglyceride(2571-8) -maximum  & 0.0983 \\
Alanine(1742-6)  -increasing  & -0.184  & Urea(3094-0)  -decreasing  & 0.0894 \\
Glucose(2345-7) -maximum  & 0.1364  & Cholesterol.in(13457-7) -minimum  & 0.0803 \\
Aspartate(1920-8) -maximum  & 0.1264  & Chloride(2075-0)  -increasing  & -0.075 \\
Cholesterol.in(13457-7)  -latest value  & 0.1119  & Aspartate(1920-8) -minimum  & 0.0737 \\
Calcium(17861-6)  -increasing  & -0.107  & Glucose(2345-7)  -increasing  & 0.0639 \\
Glucose(2345-7)  -decreasing  & -0.105  & Bilirubin(1975-2)  -increasing  & -0.060 \\
\end{tabular} \end{center} \end{table*}

\begin{table*}[t]
\caption{Top Features from the baseline model  for  425.4  Prim cardiomyopathy NEC  }
\begin{center} \begin{tabular}{ll|ll}
\multicolumn{1}{c}{Feature}&\multicolumn{1}{c}{ weight }&\multicolumn{1}{c}{Feature}&\multicolumn{1}{c}{ weight }\\ \hline 
Glucose(2345-7)  -decreasing  & -0.272  & Urea(3094-0)  -increasing  & -0.010 \\
Creatinine(2160-0) -maximum  & 0.1168  & Protein(2885-2)  -latest value  & 0.0038 \\
Albumin(1751-7) -maximum  & 0.1132  & Sodium(2951-2)  -latest value  & 0.0 \\
Urea(3094-0)  -latest value  & -0.070  & Chloride(2075-0) -maximum  & 0.0 \\
Creatinine(2160-0) -minimum  & 0.0647  & Chloride(2075-0) -minimum  & 0.0 \\
Creatinine(2160-0)  -decreasing  & 0.0560  & Sodium(2951-2)  -decreasing  & 0.0 \\
Urea(3094-0) -maximum  & -0.025  & Sodium(2951-2)  -increasing  & 0.0 \\
Glucose(2345-7) -maximum  & 0.0167  & Sodium(2951-2) -maximum  & 0.0 \\
Aspartate(1920-8) -maximum  & 0.0165  & Chloride(2075-0)  -increasing  & 0.0 \\
Glucose(2345-7)  -latest value  & -0.014  & Calcium(17861-6)  -latest value  & 0.0 \\
\end{tabular} \end{center} \end{table*}

\begin{table*}[t]
\caption{Top Features from the baseline model  for  728.87 Muscle weakness-general  }
\begin{center} \begin{tabular}{ll|ll}
\multicolumn{1}{c}{Feature}&\multicolumn{1}{c}{ weight }&\multicolumn{1}{c}{Feature}&\multicolumn{1}{c}{ weight }\\ \hline 
Alanine(1742-6)  -increasing  & 0.2943  & Glucose(2345-7)  -latest value  & -0.087 \\
Creatinine(2160-0) -minimum  & 0.2361  & Calcium(17861-6)  -latest value  & -0.083 \\
Glucose(2345-7)  -decreasing  & -0.189  & Calcium(17861-6) -maximum  & -0.079 \\
Creatinine(2160-0)  -decreasing  & 0.1719  & Urea(3094-0)  -latest value  & -0.076 \\
Creatinine(2160-0)  -increasing  & 0.1644  & Alanine(1742-6) -maximum  & -0.075 \\
Creatinine(2160-0) -maximum  & 0.1479  & Alanine(1742-6)  -latest value  & -0.071 \\
Alanine(1742-6) -minimum  & -0.112  & Albumin(1751-7) -minimum  & 0.0697 \\
Triglyceride(2571-8) -maximum  & -0.108  & Cholesterol.in(13457-7)  -increasing  & -0.067 \\
Urea(3094-0) -minimum  & 0.1040  & Protein(2885-2)  -decreasing  & -0.064 \\
Aspartate(1920-8) -minimum  & 0.1020  & Protein(2885-2)  -increasing  & 0.0623 \\
\end{tabular} \end{center} \end{table*}
\clearpage

\begin{table*}[t]
\caption{Top Features from the baseline model  for  620.2  Ovarian cyst NEC/NOS  }
\begin{center} \begin{tabular}{ll|ll}
\multicolumn{1}{c}{Feature}&\multicolumn{1}{c}{ weight }&\multicolumn{1}{c}{Feature}&\multicolumn{1}{c}{ weight }\\ \hline 
Alanine(1742-6)  -decreasing  & 0.5072  & Protein(2885-2)  -decreasing  & 0.1532 \\
Cholesterol.in(13457-7) -minimum  & -0.287  & Urea(3094-0) -minimum  & -0.147 \\
Aspartate(1920-8)  -decreasing  & -0.285  & Cholesterol.in(13457-7)  -increasing  & -0.136 \\
Alanine(1742-6)  -latest value  & -0.232  & Alanine(1742-6) -maximum  & -0.131 \\
Urea(3094-0)  -decreasing  & -0.213  & Glucose(2345-7)  -decreasing  & 0.1079 \\
Creatinine(2160-0)  -latest value  & 0.2003  & Aspartate(1920-8)  -latest value  & 0.1043 \\
Chloride(2075-0) -maximum  & 0.1717  & Glucose(2345-7) -minimum  & 0.1004 \\
Urea(3094-0) -maximum  & 0.1681  & Albumin(1751-7) -maximum  & 0.0953 \\
Creatinine(2160-0) -maximum  & -0.157  & Urea(3097-3)  -latest value  & 0.0884 \\
Glucose(2345-7)  -increasing  & -0.156  & Albumin(1751-7)  -latest value  & 0.0792 \\
\end{tabular} \end{center} \end{table*}

\begin{table*}[t]
\caption{Top Features from the baseline model  for  286.9  Coagulat defect NEC/NOS  }
\begin{center} \begin{tabular}{ll|ll}
\multicolumn{1}{c}{Feature}&\multicolumn{1}{c}{ weight }&\multicolumn{1}{c}{Feature}&\multicolumn{1}{c}{ weight }\\ \hline 
Chloride(2075-0) -minimum  & -0.337  & Creatinine(2160-0) -maximum  & 0.1148 \\
Carbon(2028-9)  -increasing  & 0.3219  & Glucose(2345-7) -maximum  & 0.1124 \\
Glucose(2345-7)  -decreasing  & -0.300  & Creatinine(2160-0)  -latest value  & 0.1008 \\
Creatinine(2160-0) -minimum  & 0.2578  & Aspartate(1920-8)  -decreasing  & 0.0993 \\
Triglyceride(2571-8)  -increasing  & -0.203  & Albumin(1751-7)  -latest value  & -0.090 \\
Albumin(1751-7) -maximum  & 0.1802  & Creatinine(2160-0)  -increasing  & 0.0865 \\
Alanine(1742-6) -minimum  & -0.179  & Sodium(2951-2) -maximum  & -0.082 \\
Cholesterol(2093-3)  -latest value  & -0.147  & Urea(3097-3)  -latest value  & -0.080 \\
Protein(2885-2)  -decreasing  & -0.139  & Alanine(1742-6)  -increasing  & 0.0764 \\
Triglyceride(2571-8)  -latest value  & -0.123  & Potassium(2823-3)  -latest value  & -0.075 \\
\end{tabular} \end{center} \end{table*}

\begin{table*}[t]
\caption{Top Features from the baseline model  for  443.9  Periph vascular dis NOS  }
\begin{center} \begin{tabular}{ll|ll}
\multicolumn{1}{c}{Feature}&\multicolumn{1}{c}{ weight }&\multicolumn{1}{c}{Feature}&\multicolumn{1}{c}{ weight }\\ \hline 
Glucose(2345-7)  -decreasing  & -0.395  & Creatinine(2160-0) -maximum  & 0.1184 \\
Alanine(1742-6)  -increasing  & 0.2594  & Triglyceride(2571-8) -minimum  & 0.1182 \\
Alanine(1742-6) -minimum  & -0.227  & Triglyceride(2571-8)  -increasing  & -0.116 \\
Cholesterol.in(13457-7)  -latest value  & -0.199  & Sodium(2951-2)  -increasing  & 0.1053 \\
Aspartate(1920-8)  -latest value  & 0.1825  & Chloride(2075-0) -minimum  & -0.095 \\
Creatinine(2160-0) -minimum  & 0.1554  & Urea(3094-0)  -increasing  & 0.0914 \\
Glucose(2345-7) -maximum  & 0.1403  & Albumin/Globulin(1759-0) -maximum  & -0.077 \\
Protein(2885-2)  -decreasing  & -0.133  & Aspartate(1920-8)  -increasing  & -0.075 \\
Urea(3094-0)  -latest value  & -0.128  & Aspartate(1920-8) -minimum  & 0.0724 \\
Glucose(2345-7) -minimum  & -0.123  & Cholesterol.in(13457-7)  -decreasing  & -0.065 \\
\end{tabular} \end{center} \end{table*}
\clearpage

\begin{table*}[t]
\caption{Top Features from the baseline model  for  362.51 Nonexudat macular degen  }
\begin{center} \begin{tabular}{ll|ll}
\multicolumn{1}{c}{Feature}&\multicolumn{1}{c}{ weight }&\multicolumn{1}{c}{Feature}&\multicolumn{1}{c}{ weight }\\ \hline 
Creatinine(2160-0)  -latest value  & -0.117  & Sodium(2951-2)  -decreasing  & 0.0 \\
Glucose(2345-7)  -decreasing  & -0.080  & Sodium(2951-2)  -increasing  & 0.0 \\
Protein(2885-2)  -latest value  & -0.060  & Sodium(2951-2) -minimum  & 0.0 \\
Creatinine(2160-0)  -increasing  & 0.0179  & Sodium(2951-2) -maximum  & 0.0 \\
Glucose(2345-7)  -latest value  & -0.015  & Calcium(17861-6)  -decreasing  & 0.0 \\
Creatinine(2160-0) -maximum  & 0.0099  & Chloride(2075-0) -minimum  & 0.0 \\
Glucose(2345-7) -minimum  & 0.0075  & Calcium(17861-6)  -increasing  & 0.0 \\
Urea(3094-0)  -latest value  & -0.001  & Calcium(17861-6) -minimum  & 0.0 \\
Chloride(2075-0) -maximum  & 0.0  & Calcium(17861-6) -maximum  & 0.0 \\
Sodium(2951-2)  -latest value  & 0.0  & Cholesterol.in(13457-7)  -latest value  & 0.0 \\
\end{tabular} \end{center} \end{table*}

\begin{table*}[t]
\caption{Top Features from the baseline model  for  414.9  Chr ischemic hrt dis NOS  }
\begin{center} \begin{tabular}{ll|ll}
\multicolumn{1}{c}{Feature}&\multicolumn{1}{c}{ weight }&\multicolumn{1}{c}{Feature}&\multicolumn{1}{c}{ weight }\\ \hline 
Glucose(2345-7)  -decreasing  & -0.328  & Chloride(2075-0)  -latest value  & 0.1241 \\
Glucose(2345-7) -maximum  & 0.2678  & Alanine(1742-6) -minimum  & -0.116 \\
Aspartate(1920-8)  -increasing  & -0.264  & Sodium(2951-2)  -increasing  & 0.1142 \\
Cholesterol.in(13457-7)  -latest value  & -0.229  & Cholesterol.in(13457-7)  -increasing  & -0.108 \\
Creatinine(2160-0)  -decreasing  & 0.1837  & Cholesterol(2093-3) -minimum  & -0.103 \\
Creatinine(2160-0) -minimum  & 0.1787  & Triglyceride(2571-8)  -increasing  & -0.096 \\
Aspartate(1920-8)  -latest value  & 0.1648  & Chloride(2075-0)  -decreasing  & -0.070 \\
Alanine(1742-6)  -decreasing  & 0.1462  & Urea(3094-0) -maximum  & 0.0699 \\
Creatinine(2160-0)  -latest value  & -0.139  & Calcium(17861-6) -maximum  & -0.069 \\
Cholesterol(2093-3)  -latest value  & 0.1360  & Potassium(2823-3)  -decreasing  & 0.0680 \\
\end{tabular} \end{center} \end{table*}

\begin{table*}[t]
\caption{Top Features from the baseline model  for  781.2  Abnormality of gait  }
\begin{center} \begin{tabular}{ll|ll}
\multicolumn{1}{c}{Feature}&\multicolumn{1}{c}{ weight }&\multicolumn{1}{c}{Feature}&\multicolumn{1}{c}{ weight }\\ \hline 
Alanine(1742-6)  -increasing  & 0.1921  & Cholesterol.in(13457-7) -minimum  & 0.1334 \\
Glucose(2345-7)  -latest value  & -0.180  & Alanine(1742-6) -maximum  & -0.128 \\
Creatinine(2160-0)  -increasing  & 0.1772  & Aspartate(1920-8)  -latest value  & 0.1276 \\
Triglyceride(2571-8)  -increasing  & -0.170  & Chloride(2075-0)  -latest value  & 0.0991 \\
Creatinine(2160-0) -minimum  & 0.1568  & Creatinine(2160-0)  -latest value  & -0.097 \\
Calcium(17861-6)  -decreasing  & 0.1538  & Aspartate(1920-8)  -increasing  & 0.0878 \\
Chloride(2075-0) -maximum  & -0.148  & Alanine(1742-6)  -latest value  & -0.081 \\
Creatinine(2160-0)  -decreasing  & 0.1446  & Cholesterol.in(13457-7)  -increasing  & -0.077 \\
Creatinine(2160-0) -maximum  & 0.1418  & Glucose(2345-7)  -decreasing  & -0.074 \\
Protein(2885-2)  -decreasing  & -0.141  & Potassium(2823-3)  -latest value  & -0.069 \\
\end{tabular} \end{center} \end{table*}
\clearpage

\begin{table*}[t]
\caption{Top Features from the baseline model  for  280.9  Iron defic anemia NOS  }
\begin{center} \begin{tabular}{ll|ll}
\multicolumn{1}{c}{Feature}&\multicolumn{1}{c}{ weight }&\multicolumn{1}{c}{Feature}&\multicolumn{1}{c}{ weight }\\ \hline 
Glucose(2345-7)  -decreasing  & -0.360  & Urea(3094-0) -minimum  & 0.1416 \\
Urea(3094-0)  -increasing  & 0.2385  & Triglyceride(2571-8) -minimum  & -0.141 \\
Carbon(2028-9)  -increasing  & 0.2151  & Protein(2885-2) -minimum  & -0.127 \\
Cholesterol.in(13457-7)  -latest value  & -0.208  & Carbon(2028-9) -maximum  & -0.125 \\
Cholesterol.in(13457-7)  -increasing  & -0.205  & Creatinine(2160-0) -minimum  & 0.1219 \\
Aspartate(1920-8)  -decreasing  & 0.1938  & Aspartate(1920-8)  -increasing  & 0.1147 \\
Creatinine(2160-0)  -increasing  & 0.1650  & Protein(2885-2)  -latest value  & 0.1064 \\
Alanine(1742-6)  -increasing  & 0.1560  & Potassium(2823-3) -maximum  & 0.1056 \\
Chloride(2075-0)  -latest value  & 0.1500  & Urea(3097-3) -minimum  & -0.097 \\
Creatinine(2160-0)  -decreasing  & 0.1487  & Creatinine(2160-0) -maximum  & 0.0954 \\
\end{tabular} \end{center} \end{table*}

%% file: cameraready.bbl
\begin{thebibliography}{32}
\providecommand{\natexlab}[1]{#1}
\providecommand{\url}[1]{\texttt{#1}}
\expandafter\ifx\csname urlstyle\endcsname\relax
  \providecommand{\doi}[1]{doi: #1}\else
  \providecommand{\doi}{doi: \begingroup \urlstyle{rm}\Url}\fi

\bibitem[Bergstra and Bengio(2012)]{bergstra2012}
J~Bergstra and Y~Bengio.
\newblock Random search for hyper-parameter optimization.
\newblock \emph{Journal of Machine Learning Research}, 2012.

\bibitem[Che et~al.(2015)Che, Kale, Li, Bahadori, and Liu]{che2015deep}
Zhengping Che, David Kale, Wenzhe Li, Mohammad~Taha Bahadori, and Yan Liu.
\newblock Deep computational phenotyping.
\newblock In \emph{Proceedings of the 21th ACM SIGKDD International Conference
  on Knowledge Discovery and Data Mining}, pages 507--516. ACM, 2015.

\bibitem[Choi et~al.(2015)Choi, Bahadori, and Sun]{choi2015doctor}
Edward Choi, Mohammad~Taha Bahadori, and Jimeng Sun.
\newblock Doctor ai: Predicting clinical events via recurrent neural networks.
\newblock \emph{arXiv preprint arXiv:1511.05942}, 2015.

\bibitem[Collobert et~al.(2011)Collobert, Kavukcuoglu, and
  Farabet]{collobert2011torch7}
Ronan Collobert, Koray Kavukcuoglu, and Cl{\'e}ment Farabet.
\newblock Torch7: A matlab-like environment for machine learning.
\newblock In \emph{BigLearn, NIPS Workshop}, number EPFL-CONF-192376, 2011.

\bibitem[Echouffo-Tcheugui and Kengne(2012)]{riskmodelreview}
JB~Echouffo-Tcheugui and AP~Kengne.
\newblock Risk models to predict chronic kidney disease and its progression: a
  systematic review.
\newblock \emph{PLoS Medicine}, 2012.

\bibitem[Firat et~al.(2016)Firat, Cho, and Bengio]{firat2016multi}
Orhan Firat, Kyunghyun Cho, and Yoshua Bengio.
\newblock Multi-way, multilingual neural machine translation with a shared
  attention mechanism.
\newblock \emph{arXiv preprint arXiv:1601.01073}, 2016.

\bibitem[Fraccaro et~al.(2016)Fraccaro, van~der Veer, Brown, Prosperi,
  O'Donoghue, Collins, Buchan, and Peek]{fraccaro2016}
Pablo Fraccaro, Sabine van~der Veer, Benjamin Brown, Mattia Prosperi, Donal
  O'Donoghue, Gary Collins, Iain Buchan, and Niels Peek.
\newblock Risk prediction for chronic kidney disease progression using
  heterogeneous electronic health record data and time series analysis.
\newblock \emph{BMC Medicine}, 2016.

\bibitem[Graves and Schmidhuber(2005)]{graves2005framewise}
Alex Graves and J{\"u}rgen Schmidhuber.
\newblock Framewise phoneme classification with bidirectional lstm and other
  neural network architectures.
\newblock volume~18, pages 602--610. Elsevier, 2005.

\bibitem[Green et~al.(2012)Green, Ritchie, New, and P]{green2012}
D~Green, J~Ritchie, D~New, and Kalra P.
\newblock How accurately do nephrologists predict the need for dialysis within
  one year?
\newblock \emph{Nephron Clin Practice}, 2012.

\bibitem[Hagar et~al.(2014)Hagar, Albers, Pivovarov, Chase, Dukic, and
  Elhadad]{hagar2014}
Y~Hagar, DJ~Albers, R~Pivovarov, HS~Chase, V~Dukic, and N~Elhadad.
\newblock Survival analysis adapted for electronic health record data:
  Experiments with chronic kidney disease.
\newblock \emph{Statistical Analysis and Data Mining}, 2014.

\bibitem[HealthPartners-Kidney-Health-Clinic(2011)]{healthpartners2011}
HealthPartners-Kidney-Health-Clinic.
\newblock Medications commonly used in chronic kidney disease.
\newblock
  \url{https://www.healthpartners.com/ucm/groups/public/@hp/@public/documents/documents/cntrb_010921.pdf},
  2011.
\newblock Accessed: 8/1/16.

\bibitem[Hochreiter and Schmidhuber(1997)]{hochreiter1997long}
Sepp Hochreiter and J{\"u}rgen Schmidhuber.
\newblock Long short-term memory.
\newblock \emph{Neural computation}, 9\penalty0 (8):\penalty0 1735--1780, 1997.

\bibitem[Ioffe and Szegedy(2015)]{ioffe2015batch}
Sergey Ioffe and Christian Szegedy.
\newblock Batch normalization: Accelerating deep network training by reducing
  internal covariate shift.
\newblock \emph{arXiv preprint arXiv:1502.03167}, 2015.

\bibitem[KDIGO(2012)]{kdigo2012}
KDIGO.
\newblock Kdigo 2012 clinical practice guideline for the evaluation and
  management of chronic kidney disease.
\newblock
  \url{http://www.kdigo.org/clinical_practice_guidelines/pdf/CKD/KDIGO_2012_CKD_GL.pdf},
  2012.
\newblock Accessed: 7/31/16.

\bibitem[Kim(2014)]{kimconvolutional}
Yoon Kim.
\newblock Convolutional neural networks for sentence classification.
\newblock \emph{EMNLP}, 2014.

\bibitem[Kim et~al.(2016)Kim, Jernite, Sontag, and Rush]{kim2016character}
Yoon Kim, Yacine Jernite, David Sontag, and Alexander~M Rush.
\newblock Character-aware neural language models.
\newblock In \emph{Thirtieth AAAI Conference on Artificial Intelligence}, 2016.

\bibitem[Krizhevsky et~al.(2012)Krizhevsky, Sutskever, and
  Hinton]{krizhevsky2012imagenet}
Alex Krizhevsky, Ilya Sutskever, and Geoffrey~E Hinton.
\newblock Imagenet classification with deep convolutional neural networks.
\newblock In \emph{Advances in neural information processing systems}, pages
  1097--1105, 2012.

\bibitem[Lasko et~al.(2013)Lasko, Denny, and Levy]{lasko2013computational}
Thomas~A Lasko, Joshua~C Denny, and Mia~A Levy.
\newblock Computational phenotype discovery using unsupervised feature learning
  over noisy, sparse, and irregular clinical data.
\newblock volume~8, page e66341. Public Library of Science, 2013.

\bibitem[Le~Cun et~al.(1990)Le~Cun, Denker, Henderson, Howard, Hubbard, and
  Jackel]{le1990handwritten}
B~Boser Le~Cun, John~S Denker, D~Henderson, Richard~E Howard, W~Hubbard, and
  Lawrence~D Jackel.
\newblock Handwritten digit recognition with a back-propagation network.
\newblock In \emph{Advances in neural information processing systems}.
  Citeseer, 1990.

\bibitem[LeCun et~al.(1998)LeCun, Bottou, Bengio, and
  Haffner]{lecun1998gradient}
Yann LeCun, L{\'e}on Bottou, Yoshua Bengio, and Patrick Haffner.
\newblock Gradient-based learning applied to document recognition.
\newblock volume~86, pages 2278--2324. IEEE, 1998.

\bibitem[Ling et~al.(2015)Ling, Trancoso, Dyer, and Black]{ling2015character}
Wang Ling, Isabel Trancoso, Chris Dyer, and Alan~W Black.
\newblock Character-based neural machine translation.
\newblock \emph{arXiv preprint arXiv:1511.04586}, 2015.

\bibitem[Lipton et~al.(2015)Lipton, Kale, Elkan, and
  Wetzell]{lipton2015learning}
Zachary~C Lipton, David~C Kale, Charles Elkan, and Randall Wetzell.
\newblock Learning to diagnose with lstm recurrent neural networks.
\newblock \emph{arXiv preprint arXiv:1511.03677}, 2015.

\bibitem[Mikolov et~al.(2013)Mikolov, Sutskever, Chen, Corrado, and
  Dean]{mikolov2013distributed}
Tomas Mikolov, Ilya Sutskever, Kai Chen, Greg~S Corrado, and Jeff Dean.
\newblock Distributed representations of words and phrases and their
  compositionality.
\newblock In \emph{Advances in neural information processing systems}, pages
  3111--3119, 2013.

\bibitem[Nair and Hinton(2010)]{nair2010rectified}
Vinod Nair and Geoffrey~E Hinton.
\newblock Rectified linear units improve restricted boltzmann machines.
\newblock In \emph{Proceedings of the 27th International Conference on Machine
  Learning (ICML-10)}, pages 807--814, 2010.

\bibitem[NICE(2014)]{nice2014}
NICE.
\newblock Chronic kidney disease in adults: assessment and management.
\newblock \url{https://www.nice.org.uk/guidance/cg182/}, 2014.
\newblock Accessed: 8/1/16.

\bibitem[Perotte et~al.(2015)Perotte, Ranganath, Hirsch, Blei, and
  Elhadad]{perotte2015}
A~Perotte, R~Ranganath, JS~Hirsch, D~Blei, and N~Elhadad.
\newblock Risk prediction for chronic kidney disease progression using
  heterogeneous electronic health record data and time series analysis.
\newblock \emph{JAMIA}, 2015.

\bibitem[Razavian and Sontag(2015)]{razavian2015temporal}
Narges Razavian and David Sontag.
\newblock Temporal convolutional neural networks for diagnosis from lab tests.
\newblock \emph{arXiv:1511.07938}, 2015.

\bibitem[Srivastava et~al.(2014)Srivastava, Hinton, Krizhevsky, Sutskever, and
  Salakhutdinov]{srivastava2014dropout}
Nitish Srivastava, Geoffrey Hinton, Alex Krizhevsky, Ilya Sutskever, and Ruslan
  Salakhutdinov.
\newblock Dropout: A simple way to prevent neural networks from overfitting.
\newblock volume~15, pages 1929--1958. JMLR, 2014.

\bibitem[Tangri et~al.(2011)Tangri, Stevens, Griffith, Tighiouart, Djurdjev,
  Naimark, Levin, and Levey]{tangri2011}
N~Tangri, L~Stevens, J~Griffith, H~Tighiouart, O~Djurdjev, D~Naimark, A~Levin,
  and A~Levey.
\newblock A predictive model for progression of chronic kidney disease to
  kidney failure.
\newblock \emph{JAMA}, 2011.

\bibitem[UK-Renal-Assocation(2014)]{ukrenalassociation2014}
UK-Renal-Assocation.
\newblock Uk renal assocation planning, initiating and withdrawal of renal
  replacement therapy.
\newblock
  \url{http://www.renal.org/guidelines/modules/planning-initiating-and-withdrawal-of-renal-replacement-therapy#sthash.y62zbp1w.hwwtAiRh.dpbs},
  2014.
\newblock Accessed: 8/1/16.

\bibitem[Zeiler(2012)]{zeiler2012adadelta}
Matthew~D Zeiler.
\newblock Adadelta: an adaptive learning rate method.
\newblock \emph{arXiv preprint arXiv:1212.5701}, 2012.

\bibitem[Zhang et~al.(2015)Zhang, Zhao, and LeCun]{zhang2015character}
Xiang Zhang, Junbo Zhao, and Yann LeCun.
\newblock Character-level convolutional networks for text classification.
\newblock In \emph{Advances in Neural Information Processing Systems}, pages
  649--657, 2015.

\end{thebibliography}
